%% file: main.tex
\documentclass[10pt,twocolumn,letterpaper]{article}

\usepackage[pagenumbers]{wacv} % To force page numbers, e.g. for an arXiv version

\input{preamble}

\definecolor{wacvblue}{rgb}{0.21,0.49,0.74}
\usepackage[pagebackref,breaklinks,colorlinks,allcolors=wacvblue]{hyperref}

\def\confName{WACV}
\def\confYear{2027}

\title{AgenTeeth: A Model-Agnostic Framework for Suppressing Hallucination in Frozen Vision-Language Models on Dental X-Rays via Tool Evidence Injection}

\author{
Ahmed Rafid\textsuperscript{1}\quad
Fariya Ahmed\textsuperscript{1,\textdagger}\quad
Rumman Adib\textsuperscript{1,\textdagger}\\
Mehedi Ahamed\textsuperscript{2}\quad
Ajwad Abrar\textsuperscript{1}\quad
Tareque Mohmud Chowdhury\textsuperscript{1}\\[4pt]
\textsuperscript{1}Islamic University of Technology, Dhaka, Bangladesh\\
\textsuperscript{2}Southeast University, Dhaka, Bangladesh\\[2pt]
\resizebox{0.95\linewidth}{!}{\texttt{\{ahmedrafid, fariyaahmed, rummanadib, ajwadabrar, tareque\}@iut-dhaka.edu}}\\
\texttt{mehedi.ahamed@seu.edu.bd}\\[2pt]
{\small\textsuperscript{\textdagger}Equal contribution as co-second authors.}
}

\begin{document}
\maketitle
\input{sec/0_abstract}    
\input{sec/Teaser_fig}
\input{sec/1_intro}
\input{sec/2_relatedwork}
\input{sec/3_domainexpert}
% =====================================================================
%  FIGURE 2, framework overview. Spans both columns.
% =====================================================================
\begin{figure*}[t]
\centering
\includegraphics[width=\textwidth]{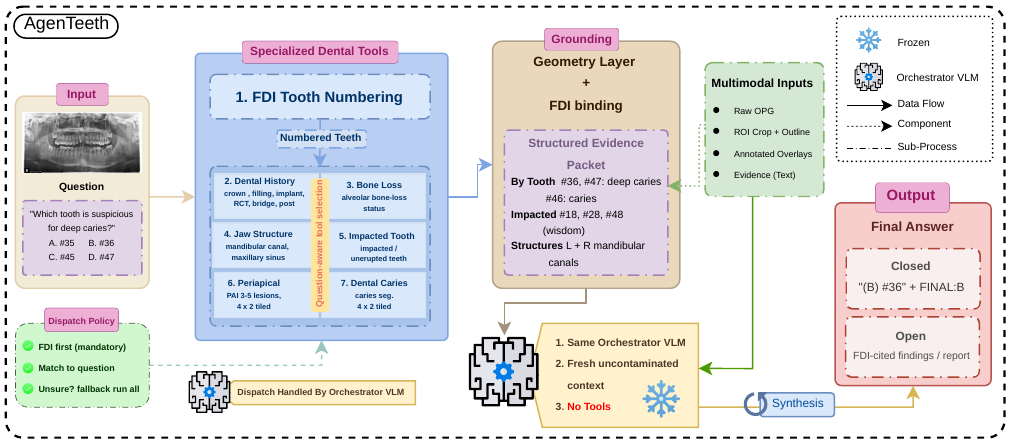}
\caption{\textbf{The AgenTeeth framework.} A frozen orchestrator VLM operates in
two passes. First, it selects the relevant domain experts, with FDI numbering
always enabled and all tools invoked when dispatch is uncertain; benchmark
category labels are never used. Detector outputs are then grounded to FDI teeth
by deterministic matching and converted into a coordinate-free evidence packet.
A fresh, tool-less second pass answers from this packet together with the raw
radiograph, selected annotated overlays, and any region crop, while prioritizing
tool evidence over unsupported visual assumptions. The matched baseline uses
the same pipeline without the seven domain experts.}
\label{fig:framework}
\end{figure*}
\input{sec/4_framework}
\input{sec/5_experimental_setup}
% =====================================================================
%  TABLES for Sec. Results.  Place near \section{Results}.
%  Both are wide; use table* (full width) in a two-column layout.
% =====================================================================

% ---------------------------------------------------------------------
% TABLE 3, main results. GPT-5-mini judge.
% ---------------------------------------------------------------------
\begin{table*}[t]
\centering
\small
\setlength{\tabcolsep}{4.2pt}
\begin{tabular}{@{}lcccccc@{\hskip 10pt}ccccccc@{}}
\toprule
& \multicolumn{6}{c}{Closed-ended VQA} & \multicolumn{7}{c}{Open-ended VQA} \\
\cmidrule(lr){2-7}\cmidrule(l){8-14}
Model & Teeth & Patho & His & Jaw & Summ & Overall
      & Teeth & Patho & His & Jaw & Summ & Report & Overall \\
\midrule
\multicolumn{14}{@{}l}{\emph{Previously reported (open split only)}}\\
GPT-5 \cite{fan2026oralgptplus}            & -- & -- & -- & -- & -- & --
  & 39.8 & 29.4 & 44.1 & \textbf{77.4} & 40.1 & 28.2 & 42.3 \\
  
OralGPT-Plus-7B \cite{fan2026oralgptplus}  & -- & -- & -- & -- & -- & --
  & 45.3 & 33.6 & 45.6 & 75.4 & 34.6 & 32.7 & 45.4 \\
\midrule
\multicolumn{14}{l}{\emph{Previously reported closed-ended result}} \\
Ovis2-34B~\cite{hao2025mmoral}
& 45.8 & 51.6 & 53.9 & 79.4 & 79.2 & 56.8
& -- & -- & -- & -- & -- & -- & -- \\
\midrule
\multicolumn{14}{@{}l}{\emph{Baseline: identical pipeline, detectors removed}}\\
gemma4:31b      & 43.7 & 35.8 & 54.9 & 76.7 & 58.1 & 54.7
                & 32.7 & 25.7 & 44.4 & 53.8 & 34.6 & 28.6 & 35.7 \\
qwen3.5         & 53.8 & 48.6 & 68.3 & 77.5 & 58.1 & 62.3
                & 48.9 & 38.9 & 61.5 & 78.9 & 39.8 & 30.4 & 49.9 \\
mistral-large-3 & 37.7 & 35.8 & 52.1 & 49.6 & 61.3 & 40.7
                & 37.7 & 37.1 & 42.7 & 52.2 & 45.4 & 26.5 & 38.8 \\
minimax-3       & 44.5 & 33.8 & 64.1 & 76.0 & 61.3 & 51.3
                & 41.9 & 28.6 & 61.2 & 67.1 & 45.1 & 29.9 & 44.5 \\
\midrule
\multicolumn{14}{@{}l}{\emph{AgenTeeth (ours)}}\\
gemma4:31b      & 66.2 & 59.5 & 71.8 & 78.3 & 67.7 & 69.0
                & 52.9 & 54.4 & 63.3 & 76.3 & 50.7 & 44.0 & 56.0 \\
qwen3.5         & \textbf{73.5} & \textbf{66.9} & 76.1 & 76.7 & \textbf{74.2} & \textbf{74.1}
                & 62.1 & 59.2 & 68.5 & 87.6 & 65.6 & \textbf{47.6} & 63.9 \\
mistral-large-3 & 60.3 & 56.1 & 67.6 & \textbf{79.1} & 71.0 & 64.4
                & 57.9 & 57.6 & 69.1 & \textbf{89.1} & 65.7 & 39.9 & 61.1 \\
minimax-3       & 69.9 & 63.5 & \textbf{76.1} & \textbf{79.1} & 71.0 & 72.1
                & \textbf{64.2} & \textbf{65.6} & \textbf{69.5} & 88.8 & \textbf{69.8} & 45.6 & \textbf{65.7} \\
\midrule
\multicolumn{14}{@{}l}{\emph{Backbone-matched: Qwen2.5-VL-7B with frozen orchestrator}}\\
Qwen2.5-VL-7B + AgenTeeth & -- & -- & -- & -- & -- & --
                & 41.9 & 53.4 & 58.4 & 59.3 & 55.6 & 39.1 & \underline{48.1} \\
\bottomrule
\end{tabular}
\caption{\textbf{MMOral-OPG-Bench results}, in percent. Open-ended results are
scored with the GPT-5-mini evaluation protocol used by OralGPT-Plus, while
closed-ended results use exact-match accuracy and are judge-independent.
Baseline rows use the identical AgenTeeth pipeline with the seven domain experts
removed. Previously reported open-ended rows are from OralGPT-Plus; the strongest
closed-ended MMOral result is included for comparison. Closed accuracy is capped
at 93.5\% by 32 unanswerable questions.}
\label{tab:main}
\end{table*}
\input{sec/6_results}
% ---------------------------------------------------------------------
% TABLE 4, judge robustness
% ---------------------------------------------------------------------
% =====================================================================
%  TABLE 3, judge robustness. Replaces the deltas-only version.
%  Every number is now shown, so the table reads without the prose.
%
%  Source: Model_Performances.xlsx, sheets judegGPT-4o and judgeGPT-5-mini.
%  Closed-ended Overall is exact-match and identical under both judges,
%  so it appears once.
% =====================================================================
% =====================================================================
\begin{table}[t]
\centering\small\setlength{\tabcolsep}{4.5pt}
\begin{tabular}{@{}l@{\hskip 7pt}ccc@{\hskip 8pt}ccc@{}}
\toprule
& \multicolumn{3}{c}{GPT-4o judge} & \multicolumn{3}{c}{GPT-5-mini judge} \\
\cmidrule(lr){2-4}\cmidrule(l){5-7}
Model & Base & Ours & $\Delta$ & Base & Ours & $\Delta$ \\
\midrule
gemma4:31b      & 34.7 & 51.8 & +17.1 & 35.7 & 56.0 & +20.3 \\
qwen3.5         & 45.1 & 60.6 & +15.5 & 49.9 & 63.9 & +14.0 \\
mistral-large-3 & 38.3 & 59.0 & +20.7 & 38.8 & 61.1 & +22.4 \\
minimax-3       & 35.9 & 55.9 & +19.9 & 44.5 & 65.7 & +21.2 \\
\bottomrule
\end{tabular}
\caption{\textbf{Open-ended MMOral-OPG-Bench Overall under two judges}, in percent.
\emph{Base} denotes the matched detector-free baseline and \emph{Ours} denotes
AgenTeeth. AgenTeeth improves all four backbones under both GPT-4o and
GPT-5-mini, showing that the performance gains are consistent across judge
choice. Closed-ended results are judge-independent and are reported separately
in Tab.~\ref{tab:main}.}
\label{tab:judge}
\end{table}
\input{sec/7_ablation_studies}
\input{sec/8_limitations}
\input{sec/9_conclusion}
{
    \small
    \bibliographystyle{ieeenat_fullname}
    \bibliography{main}
}

\end{document}

% Prompts are set single-column: system prompts exceed the width of a
% 3.4cm column and wrap unreadably otherwise. Sec. B onward reverts to
% the two-column layout of the main paper.
\onecolumn
\title{AgenTeeth: Supplementary Material}
\maketitle
\appendix

% =====================================================================
\section{Prompts}
\label{sup:prompts}
% =====================================================================

We reproduce every prompt used to produce the reported results, verbatim.
Both configurations are built from parallel prompt sets that share their
answer-format, report-structure and citation instructions; the tool-aided
set additionally describes the detector evidence, and the baseline set
omits every mention of detectors.

\subsection{Stage 1: Dispatch}
\label{sup:prompt-dispatch}

Dispatch selects which detectors to invoke from the question text alone.
The benchmark's category label is never exposed to the model. In the
tool-aided configuration the dispatcher sees the seven domain detectors;
in the baseline it sees only the region-inspection tool. Both terminate
with the same sentinel token, so the two-stage call structure is
identical.

\paragraph{Tool-aided configuration.}
\begin{prompt}
You are a dental diagnostic AI agent with access to specialised detection tools.
Decide which tools to call based ONLY on what the question asks.

AVAILABLE TOOLS
- run_fdi_tool ............ numbers every tooth (FDI). ALWAYS REQUIRED: every
                           other finding is reported against a tooth number.
- run_dental_history_tool . crowns, fillings, implants, root canals, bridges, posts
- run_bone_loss_tool ...... alveolar bone loss / periodontal status
- run_jaw_structure_tool .. maxillary sinus, mandibular canal, condyle, TMJ
- run_impacted_tooth_tool . impacted / unerupted teeth
- run_periapical_tool ..... periapical lesions, abscess, granuloma, cyst (PAI 3-5)
- run_dental_caries_tool .. caries, cavities, decay

RULES
1. run_fdi_tool is ALWAYS required. Call it first.
2. Call every tool whose topic the question touches.
3. If you are UNSURE which tool applies, call ALL of them.
4. If the question asks for a diagnosis, a full report, an overall assessment,
   a summary, or any general "what do you see" reading, you MUST call ALL tools.
5. Call each tool at most once.

IMPORTANT
- Do NOT answer the question yet. Only call tools.
- After the tool calls complete, reply with exactly: TOOLS_COMPLETE
\end{prompt}

Rule 4 is the reason the cache described in Sec.~\ref{sup:caching}
matters: every one of the 100 open-ended report questions invokes the
full detector set.

\paragraph{Baseline configuration.}
\begin{prompt}
You are a dental diagnostic AI agent reading a panoramic dental radiograph.

You have ONE tool:
- inspect_region(x1, y1, x2, y2): returns a magnified crop of that rectangle,
  in absolute pixel coordinates of the full image.

WHEN TO USE IT
- If the question points at a specific area or asks about a small feature that
  is hard to resolve at full size, inspect that area.
- If the question names a bounding box, that region has ALREADY been cropped
  for you - you do not need to request it again.
- Otherwise you do not need the tool.

IMPORTANT
- Do NOT answer the question yet.
- When you are finished inspecting (or if no inspection is needed), reply with
  exactly: TOOLS_COMPLETE
\end{prompt}

\subsection{Stage 2: Synthesis}
\label{sup:prompt-synthesis}

Synthesis is a fresh call with no tool access and no dispatch history. A
shared base prompt is concatenated with a closed-ended or open-ended
suffix, and with the region addendum when the question names a bounding
box.

\paragraph{Tool-aided base.} The evidence-primacy instruction discussed
in the main paper is the second rule. The remaining rules are the
guardrails that constrain it: unbound findings must not be given a tooth
number, tool errors must be surfaced rather than guessed around, and
silence from a detector is not evidence of absence.

\begin{prompt}
You are a dental diagnostic AI agent producing a final answer.

You have:
1. The original panoramic X-ray (first image)
2. Annotated X-rays from the detection tools (subsequent images)
3. Structured tool evidence below, already bound to FDI tooth numbers

RULES
- Ground every claim in the tool evidence and name the tool that supports it.
- If the tool evidence contradicts your visual impression, trust the tools.
- If a tool errored, state the resulting uncertainty rather than guessing.
- Findings listed as UNBOUND could not be attributed to a specific tooth; do
  not invent a tooth number for them.
- Absence of a detection is not proof of absence - say so when it matters.
\end{prompt}

\paragraph{Baseline base.} The same role and the same output discipline,
with the evidence rules replaced by an instruction to read the
radiograph directly.

\begin{prompt}
You are a dental diagnostic AI agent producing a final answer.

You have:
1. The original panoramic X-ray (first image)
2. Any zoomed region crops (subsequent images, if present)

RULES
- Base your answer on your own reading of the radiograph.
- Be specific and concise.
- Where the image does not support a confident call, say so rather than
  inventing detail.
\end{prompt}

\paragraph{Open-ended suffix.} Both configurations require findings to be
cited by FDI number. The tool-aided variant adds one clause discouraging
unsupported hedging.

\begin{prompt}
This is an open-ended question. Answer in clear clinical prose.

- Cite the exact FDI tooth number for every diagnosis you state.
- Be specific and concise; do not pad with caveats the evidence does not
  support.                                            [tool-aided only]
\end{prompt}

\paragraph{Region addendum.} Appended when the question names a bounding
box. The tool-aided variant receives a third image outlining the region
on the full radiograph and a rule for resolving overlapping structures.

\begin{prompt}
[tool-aided]
A specific REGION OF INTEREST was given in the question. The second image is a
zoomed crop of that region and the third outlines it on the full radiograph.
Several structures may overlap the box - prefer the one with the highest IoU in
REGION FINDINGS.

[baseline]
A specific REGION OF INTEREST was named in the question. The second image is a
zoomed crop of that region. Answer about THAT region specifically.
\end{prompt}

\subsection{Conditional Caveat Blocks}
\label{sup:prompt-caveats}

Two blocks are appended conditionally, each encoding a documented failure
mode of a specific detector rather than a general instruction to be
cautious. Both are present in the tool-aided configuration only.

\paragraph{Anatomical plausibility.} Appended when the periapical or
caries detector ran. These two detectors produce a characteristic class
of anatomically impossible false positive, and this block licenses the
model to discard them. It is the one point at which evidence primacy is
explicitly overridden, and it is overridden by anatomy rather than by the
model's visual impression.

\begin{prompt}
ANATOMICAL PLAUSIBILITY CHECK (apply before trusting these detections):
- A PERIAPICAL lesion can ONLY occur at the root apex of a tooth, inside the
  tooth-bearing (alveolar) bone of the upper or lower jaw. A periapical
  detection located in the upper skull, the nasal / sinus / orbital area, or
  anywhere that is NOT at a tooth root apex is a FALSE POSITIVE - ignore it. In
  the annotated periapical image, disregard any box sitting high above the
  tooth roots. Any periapical finding listed as UNBOUND (not attributable to a
  tooth) should be treated as spurious unless you can clearly see it at a real
  root apex.
- DENTAL CARIES can ONLY occur on a tooth. A caries detection over jaw bone
  where no tooth is present (an edentulous gap, the ramus or condyle, or empty
  bone) is a FALSE POSITIVE - ignore it. Any caries finding listed as UNBOUND
  should be treated as spurious.
Base your answer only on detections that are anatomically plausible; do not
count these false positives toward lesion or caries totals.
\end{prompt}

\paragraph{Jaw structure.} Appended when the jaw-structure detector ran.
That detector has no nasal-cavity class, so its silence about the nasal
cavity carries no information, and the block redirects the model to the
raw radiograph using the maxillary sinuses as landmarks.

\begin{prompt}
NASAL CAVITY (the jaw-structure tool cannot detect it):
- The jaw-structure detector reports ONLY mandibular canals and maxillary
  sinuses. It has no nasal-cavity class, so it will NEVER list a nasal cavity
  even when one is clearly visible. Do NOT conclude the nasal cavity is absent
  or not visible just because the tool evidence omits it.
- To judge the nasal cavity, look at the ORIGINAL X-ray, not the tool list. The
  nasal cavity sits in the midline, BETWEEN and slightly above the two maxillary
  sinuses. When both maxillary sinuses are detected, inspect the midline
  radiolucent air space between them: if that central space above the upper
  front teeth is visible, the nasal cavity is visible.
\end{prompt}

\subsection{Report Structure Prompt}
\label{sup:prompt-report}

Report-style questions are answered under a fixed structure derived from
the components MMOral-OPG-Bench identifies as expected in a written
dental report. \textbf{This prompt is identical in both configurations.}
It is a confound control rather than part of the tool condition: without
it, differences in report score would partly reflect formatting rather
than content.

\begin{prompt}
This question asks for a full radiographic REPORT / caption of findings for the
whole image. Write your answer using these exact markdown sections, including a
line only where there is an actual finding:

### Teeth-Specific Observations
1. General Condition: the total number of teeth visualized, and the wisdom teeth
   (#18, #28, #38, #48) that are present with their status (erupted or impacted).
2. Pathological Findings: each affected tooth by FDI number with its caries or
   periapical finding (e.g. "#46: imaging features sign of periapical lesion").
3. Historical Interventions: each tooth by FDI number with its restoration
   (crown, filling, root canal treatment, bridge, implant, or post).

### Jaw-Specific Observations
1. Bone Architecture: any alveolar bone loss, or "No apparent bone loss" if none.
2. Visible Structures: visible non-tooth structures (mandibular canals,
   maxillary sinuses, etc.).

### Clinical Summary & Recommendations
1. Priority Concerns: the findings that need urgent attention.
2. Preventive measures or follow-up recommendations where appropriate.

Cite the exact FDI tooth number for every diagnosis. Only include findings you
can support; omit a section that has none.
\end{prompt}

\subsection{Answer Format}
\label{sup:prompt-format}

Closed-ended answers terminate in a fixed marker consumed by the
exact-match parser. \textbf{This block is byte-identical in both
configurations}, which is required rather than incidental: the official
MMOral evaluator searches for the parenthesised letter first and falls
back to a random guess when it finds none, so any asymmetry in output
formatting between the two configurations would appear as a difference
in accuracy.

\begin{prompt}
OUTPUT FORMAT (strict):
Begin your reply with the chosen option letter in parentheses, then one
sentence of justification. Example:
(B) The radiograph shows an impacted tooth at that site.

Then, on the final line, repeat it as:
FINAL: B

Choose ONLY from the option letters actually listed. Do not invent a letter.
\end{prompt}

The tool-aided configuration substitutes a tool-grounded example
sentence, \texttt{(B) The FDI tool detected an impacted tooth at \#38},
leaving the format specification unchanged.

\subsection{What Differs Between Configurations}
\label{sup:prompt-diff}

Collecting the above, the two configurations share the report-structure
prompt, the answer-format block, the FDI citation requirement, the
two-stage call structure with its sentinel token, the answer parser, and
the region-inspection tool. They differ in the tools exposed to the
dispatcher and, necessarily, in the synthesis rules that describe how to
use detector evidence: a baseline instructed to trust tool evidence it
does not receive would be incoherent. We therefore do not claim the two
prompt sets are textually identical, only that no instruction was given
to one configuration that could have been given to the other.

\subsection{Judge Prompt}
\label{sup:prompt-judge}

Open-ended answers are scored by an LLM judge under a few-shot rubric
producing a score in $[0,1]$. We publish this prompt because a shared
judge \emph{model} does not imply a shared scoring \emph{procedure}:
prior work reporting GPT-5-mini judge scores uses its own few-shot rubric
rather than the benchmark's stock judge prompt. Readers assessing our
cross-paper comparison should compare the two rubrics directly.

\begin{prompt}
Given the question, compare the ground truth and prediction from AI models, to generate a correctness score for the prediction.The correctness score is 0.0 (totally wrong), 0.1, 0.2, 0.3, 0.4, 0.5, 0.6, 0.7, 0.8, 0.9, or 1.0 (totally right).
Just complete the last space of the correctness score.
| Question | Ground truth | Prediction | Correctness |
| --- | --- | --- | --- |
| How many teeth are visualized in the radiograph? | 30 teeth are visualized with clear anatomical definition. | 30 | 1.0 |
| How many teeth are visualized in the radiograph? | 30 teeth are visualized with clear anatomical definition. | 29 teeth are visualized with clear anatomical definition. | 0.0 |
| What is the status of the wisdom teeth in the radiograph? | Three wisdom teeth are detected, all of which are impacted: #18, #28, and #48. | #18: impacted, #28: impacted, #48: erupted | 0.7 |
| What is the condition of the teeth #26 and #14? | Teeth #26 and #14 show signs of periapical abscesses. | Teeth #26 and #23 show signs of periapical abscesses. | 0.5 |
| What is the condition of the bone architecture and visible structures in the jaw? | No apparent bone loss is observed. Bilateral mandibular canals and maxillary sinuses are clearly visible. | Bilateral mandibular canals and maxillary sinuses are clearly visible. | 0.5 |
| What is the clinical priority concern regarding the periapical lesions? | Periapical cysts at #11 and #12, and granuloma at #46 require endodontic evaluation. | Periapical lesions at #11, #12, and #46 require endodontic evaluation. | 0.8 |
| What radiographic features are visible in tooth #31 on the panoramic X-ray? | [\n{"Teeth position": {\"point_2d\": [1242, 726]}},\n{"Crown": {\"box_2d\": [1220, 637, 1266, 741]}}\n] | Crown | 0.8 |
| What radiographic features are visible in tooth #31 on the panoramic X-ray? | [\n{"Teeth position": {\"point_2d\": [1242, 726]}},\n{"Crown": {\"box_2d\": [1220, 637, 1266, 741]}}\n] | Crown at position: [1230, 627, 1276, 750] | 0.9 |
| What radiographic features are visible in tooth #31 on the panoramic X-ray? | [\n{"Teeth position": {\"point_2d\": [1242, 726]}},\n{"Crown": {\"box_2d\": [1220, 637, 1266, 741]}}\n] | Teeth at position: {\"point_2d\": [1242, 726]}},\n{Crown at position: {\"box_2d\": [1230, 627, 1276, 750]}} | 1.0 |
\end{prompt}

\twocolumn

% =====================================================================
\section{Experimental Setup}
\label{sup:setup}
% =====================================================================

\subsection{Benchmark}

We evaluate on MMOral-OPG-Bench \cite{hao2025mmoral}, which comprises 491
closed-ended multiple-choice questions and 578 open-ended questions over
100 panoramic radiographs. Questions carry one or more of five category
labels (Teeth, Pathology, History, Jaw, Summary), with a sixth category,
Report, present in the open split only. Because labels are multi-valued,
per-category totals exceed the question count. Images are used at full
resolution, decoded from the released archive.

Two properties of the closed split bound what any system can score on it.
Thirty-two of the 491 closed questions have a reference answer pointing
at a blank option and are effectively unanswerable. Following the
benchmark protocol we retain them in the denominator, which caps closed
accuracy at $93.5\%$. The majority-class answer alone scores $44.0\%$.
Both figures should be read alongside the closed-split numbers in the
main paper.

\subsection{Controlled Comparison Design}

Our central experiment compares two configurations of the same system.
The \emph{tool-aided} configuration is AgenTeeth as described in the
Framework section of the main paper: two-stage dispatch and synthesis
with the seven domain detectors plus a region-inspection tool. The
\emph{baseline} configuration is the identical pipeline with the domain
detectors removed and the region-inspection tool retained.

Everything else is held fixed: the same base model, the same two-stage
structure and therefore the same number of model calls per question, the
same answer parser, the same report-structure instruction, and the same
decoding and rate-limit configuration. The only variable is detector
availability and the evidence derived from it, together with the
synthesis rules that necessarily accompany it
(Sec.~\ref{sup:prompt-diff}). This is what licenses a causal reading of
the difference between configurations, and it is a comparison that
designs entangling tools with training cannot make.

Three components are part of the tool condition and are disclosed here
rather than treated as neutral infrastructure: grounding coordinates
appended to open-ended answers, the anatomical-plausibility filter of
Sec.~\ref{sup:prompt-caveats}, and the two conditional caveat blocks. All
are present in the tool-aided configuration only.

\subsection{Models and Serving}

We evaluate four base VLMs, all served through Ollama Cloud:
gemma4:31b-cloud, qwen3.5-cloud, mistral-large-3 and minimax-3. To test
whether the framework holds at small scale we additionally evaluate
Qwen2.5-VL-7B-Instruct, which is the base model OralGPT-Plus
\cite{fan2026oralgptplus} fine-tunes, allowing a parameter-matched
comparison against a trained tool-using system.

Decoding used the providers' default sampling parameters; we did not
override the temperature or other generation settings. Because those
defaults apply a nonzero temperature, decoding was stochastic rather than
greedy, so a single run does not capture the resulting sampling variance.
Each configuration was nonetheless run once, without averaging over seeds
or samples, and we treat this as a limitation in the main paper.

\subsection{Tool Inference}

The seven domain detectors are computer-vision models (object detection
and instance segmentation) hosted on Roboflow and queried at run time
through its hosted Inference API \cite{roboflowdeploy}. The
region-inspection tool available to both configurations is a local image
crop and involves no hosted inference. Each detector returns structured
detections (class, confidence and geometry), which the pipeline binds to
FDI tooth numbers before they reach the synthesis stage; the base model
never sees raw coordinates.

The periapical lesion and caries detectors target small lesions and are
run tiled: the radiograph is partitioned into a $4\times2$ grid, each
cell is queried separately, and the cell-level detections are merged back
to full-image coordinates. This costs eight hosted calls per tiled
detector against a single call for each of the other five, so a full
sweep of all seven detectors over one image is roughly 21 hosted calls
($5\times1 + 2\times8$). Each detector is queried at a fixed,
tool-specific confidence threshold. Tool dispatch is driven by the
question text alone; we deliberately do not read the benchmark's category
column, since doing so would leak evaluation metadata into the system
under test.

\subsection{Caching}
\label{sup:caching}

The dispatch policy instructs the model to call every potentially
relevant detector, and to call all seven whenever the question is a
full-report or ``what do you see'' request; the 100 open-ended report
questions alone invoke the full set. With roughly eleven questions per
image, computing detectors afresh for every question would repeat the
same 21 hosted calls hundreds of times per image, at a cost in
wall-clock time and API usage that would dominate the experiment. We
therefore interpose a two-tier cache, in memory and on disk, between the
pipeline and the Inference API, keyed on the triple (image identity,
tool, confidence threshold).

The cache turns a detector invocation into a lookup. When a tool is
called, the pipeline first checks the cache for that
image--tool--threshold key. On a \emph{hit} it returns the stored
structured output directly and makes no network call. Only on a
\emph{miss}, that is when no output exists for that key, does it issue a
single hosted inference call, persist the result to memory and disk, and
then return it. Each image's detectors are therefore computed at most
once per threshold and reused across every question about that image.
Caching affects only the tool-aided configuration; the baseline calls no
hosted detector.

\subsection{Evaluation Protocol}

Closed-ended questions are scored by exact-match letter accuracy using
the official MMOral evaluation kit, over all 491 questions. Open-ended
questions are scored by an LLM judge under a few-shot rubric producing a
soft score in $[0,1]$.

Our primary judge is GPT-5-mini, matching the judge used by OralGPT-Plus
\cite{fan2026oralgptplus}. Because a shared judge model does not by
itself guarantee a shared scoring procedure, and because judge choice is
a known source of variance, we repeat the full open-split evaluation
under a second judge, GPT-4o, and report both. Closed-split scores are
judge-independent by construction.

Because benchmark categories are multi-valued, the Overall score counts
each question once rather than averaging the per-category columns.

\subsection{Threats to Validity}

Our baseline is not the zero-shot protocol used for the MMOral
leaderboard. It receives full-resolution images where the leaderboard
protocol downscales, and it retains structured prompting and the region
tool. Comparisons against published leaderboard numbers are therefore
system-versus-system, and we reserve causal claims about the contribution
of tools to the internal controlled comparison, where the confound is
removed by construction. Our base models also differ from those on the
leaderboard, so we claim that tool augmentation lifts these four systems
past previously reported results, not that it would improve any specific
leaderboard entry.

% =====================================================================
\section{Dataset Documentation}
\label{sup:datasets}
% =====================================================================

\subsection{Provenance}

Tab.~\ref{tab:sup-provenance} lists every source dataset, its citation,
and the detectors trained on it. All sources are publicly available.

\begin{table}[t]
\centering\small\setlength{\tabcolsep}{3.5pt}
\begin{tabular}{@{}p{2.9cm}p{1.5cm}p{2.7cm}@{}}
\toprule
Source & Reference & Used by \\
\midrule
Tufts Dental Database & \cite{panetta2022tufts} & FDI, bone loss, dental history \\
DUAL (Kaggle) & \cite{zhou2024dual} & FDI \\
DENTEX 2023 & \cite{hamamci2023dentex} & FDI \\
Panoramic Dental X-ray & \cite{brahmi2025pdx} & FDI, jaw structure \\
Annotated OPG (restorations) & \cite{khurshid2025opg} & Dental history \\
Panoramic radiographs with periapical lesions & \cite{do2024periapical} & Periapical \\
Multi-centre panoramic dataset & \cite{li2024multicenter} & Caries, impacted tooth \\
\bottomrule
\end{tabular}
\caption{Source datasets and the detectors trained on them.}
\label{tab:sup-provenance}
\end{table}

\subsection{Image Counts and Splits}

Tab.~\ref{tab:sup-counts} gives per-detector image counts. \emph{Images}
counts distinct source images before augmentation. For the two tiled
detectors the training unit is a tile rather than a radiograph, and both
figures are given.

\begin{table}[t]
\centering\small\setlength{\tabcolsep}{3pt}
\begin{tabular}{@{}lrrrrr@{}}
\toprule
Detector & Images & Train & Valid & Test & Aug.\ train \\
\midrule
FDI numbering    & 1{,}967 & 1{,}574 & 197 & 196 & 3{,}148 \\
Bone loss        &   943 &   755 &  94 &  94 & 1{,}510 \\
Dental history   & 2{,}456 & 1{,}935 & 261 & 260 & 3{,}870 \\
Jaw structure    &   320 &   256 &  32 &  32 &   768 \\
Impacted tooth   & 3{,}362 & 2{,}638 & 356 & 368 & 5{,}276 \\
Periapical$^\dagger$ & 3{,}744 & 3{,}006 & 364 & 374 & 6{,}012 \\
Dental caries$^\dagger$ & 2{,}470 & 1{,}977 & 248 & 245 & 3{,}954 \\
\bottomrule
\end{tabular}
\caption{Per-detector dataset sizes. $^\dagger$Counts are tiles, not
radiographs: periapical derives 3{,}744 tiles from 3{,}924 source
radiographs, and caries derives 2{,}470 tiles from 1{,}277.
\emph{Aug.\ train} is the training split after augmentation.}
\label{tab:sup-counts}
\end{table}

Dataset size varies by more than an order of magnitude across detectors,
from 320 radiographs for jaw structure to 3{,}924 for periapical lesions.
Jaw structure nonetheless achieves the highest held-out score of any
tool, which we attribute to its targets being large, high-contrast
anatomical structures rather than small lesions. Data requirements track
target difficulty rather than task type.

\subsection{Classes and Instance Counts}

\paragraph{FDI numbering.} Thirty-two classes covering permanent
dentition, eight per quadrant (\texttt{11}--\texttt{18},
\texttt{21}--\texttt{28}, \texttt{31}--\texttt{38},
\texttt{41}--\texttt{48}), with 53{,}334 instances in total. Per-class
counts range from 906 (\texttt{28}) to 1{,}938 (\texttt{43}). The
92.4 mAP@50 reported in the main paper is therefore a 32-way result, not
a single-class detection score. The project additionally contains 12
boxes labelled \texttt{91}, which is not valid FDI notation; these are an
annotation artifact and the class is not exposed at inference.

\paragraph{Bone loss.} Four severity classes with a heavily skewed
distribution: \texttt{healthy} 19{,}588, \texttt{mild} 2{,}116,
\texttt{medium} 453, \texttt{severe} 159. Healthy outnumbers severe by
123 to 1. This imbalance is the most likely explanation for the
detector's recall of $52.1\%$, discussed in the main paper's limitations.

\paragraph{Dental history.} Six classes, 13{,}950 instances:
\texttt{Dental\_Fillings} 7{,}616, \texttt{Crown} 2{,}017,
\texttt{Root\_Canal\_treated\_tooth} 1{,}731, \texttt{implant} 1{,}065,
\texttt{Endodontic\_post} 809, \texttt{bridge} 712.

\paragraph{Jaw structure.} \texttt{Mandibular Canal} 619,
\texttt{maxillary sinus} 450. As noted in
Sec.~\ref{sup:prompt-caveats}, there is no nasal-cavity class.

\paragraph{Periapical.} Three PAI severity grades: \texttt{3} 2{,}294,
\texttt{4} 1{,}777, \texttt{5} 735.

\paragraph{Dental caries.} Single class, 3{,}866 instances.

\paragraph{Impacted tooth.} The source project contains
\texttt{impacted\_tooth} (5{,}015) and \texttt{caries} (2{,}928)
annotations. The tool discards the caries class, which is served by the
dedicated caries detector.

\subsection{Annotation Protocol}
\label{sup:annotation}

Two training sets were annotated for this work: the complete bone loss
set (943 radiographs from Tufts) and approximately 200 Tufts radiographs
extending the dental history set.

\subsubsection{Annotation Protocol for Alveolar Bone Loss}

Annotations were performed by a practising dentist, who was instructed to assess
each tooth visible on the panoramic radiograph and assign it to one of four
severity categories reflecting the extent of alveolar bone loss: \textit{healthy},
\textit{mild}, \textit{medium}, or \textit{severe}. Category assignment was left
to the annotator's own clinical judgement and interpretation of the radiographic
evidence, without the imposition of fixed numerical thresholds, in order to
reflect the diagnostic reasoning applied in routine practice. For each assessed
tooth, the annotator drew a bounding box encompassing the full extent of the
tooth together with the surrounding region of alveolar bone loss associated with
it, such that both the dental structure and its adjacent bone level were
contained within the annotated region. Each bounding box was labelled with the
corresponding severity category, producing one annotation per assessed tooth.

\subsubsection{Annotation Protocol for Dental History}

Annotations for prior dental treatment were performed by a practising dentist,
who was instructed to identify every restorative or endodontic artefact visible
on the panoramic radiograph and delineate it using a polygon boundary. Five
categories of dental work were annotated:\textit{ root canal treatment, endodontic post,
crown, bridge, and filling}. Rather than tracing the artefact outline exactly, the
annotator was instructed to draw each polygon with a small margin of surrounding
headspace, so that the immediately adjacent anatomical structures were included
within the annotated region. This deliberate inclusion of peripheral context was
intended to allow the model to ground each detection in its surrounding
anatomical setting rather than relying solely on the radiopaque appearance of the
artefact itself, thereby reducing dependence on intensity cues alone. Each
polygon was labelled with its corresponding treatment category, yielding one
annotation per identified artefact.
\subsection{FDI Label Conversion}

The Tufts annotations use the universal numbering system and were
converted to FDI notation. Paediatric labels and the radiographs carrying
them were removed, so the resulting label space covers permanent
dentition only. This matters for the merged set: mixing primary and
permanent labels would introduce classes that the quadrant arithmetic of
the main paper does not cover.

\subsection{Split Policy}
\label{sup:splits}

Each source was split independently and the splits merged like for like,
train with train and validation with validation. No global re-split was
performed after merging, so no source's images can straddle a split
boundary through the merge step.

\begin{table}[t]
\centering\small\setlength{\tabcolsep}{4pt}
\begin{tabular}{@{}lrrrr@{}}
\toprule
Source & Images & Train & Valid & Test \\
\midrule
Tufts & 878 & 703 & 88 & 87 \\
DENTEX 2023 & 621 & 489 & 70 & 62 \\
DUAL (Kaggle) & 470 & 381 & 40 & 49 \\
Panoramic Dental X-ray & 53 & 37 & 5 & 4 \\
\bottomrule
\end{tabular}
\caption{Per-source splits of the merged FDI training set.}
\label{tab:sup-fdisplit}
\end{table}

We state the evidential status of Tab.~\ref{tab:sup-fdisplit} precisely.
The near-$80/10/10$ proportions within each source are \emph{consistent
with} independent splitting, but they do not prove it, since a global
merge-then-split would produce similar per-source proportions by chance.
The evidence for the policy is the build procedure itself.

For the multi-centre source the mapping is verified rather than inferred.
That dataset ships centre-wise folders which map deterministically onto
splits: the training folder to train, the first test folder to
validation, and the remaining three test folders to test. Every image
carrying a given source token lands in exactly one split. Because the
three test folders come from centres not represented in training, the
impacted tooth and caries detectors are evaluated on a genuinely
held-out multi-centre test set rather than a random split.

For the two tiled detectors, splitting is performed at the radiograph
level before tiling, so all eight tiles of a radiograph share a split and
no tile-level leakage is possible. The periapical split is stratified by
the rarest PAI grade present in each radiograph. The periapical source
ships 3{,}924 original radiographs alongside 13{,}071 pre-made augmented
copies in a separate directory; only the originals were read, so
augmented copies of one radiograph cannot appear on both sides of a
split.

\paragraph{Residual risk.} The per-source policy does not address
near-duplicate radiographs appearing in \emph{different} sources and
landing in different splits, which is a cross-source collision. We did
not run a perceptual-hash check for such pairs and therefore cannot rule
them out.

% =====================================================================
\section{Training Details}
\label{sup:training}
% =====================================================================

\subsection{Detector Configuration}

All seven detectors are RF-DETR medium models~\cite{robinson2026rfdetr}
initialised from pretrained weights, four trained for object detection
and three for instance segmentation.

All detectors were trained using Roboflow's default hyperparameters, epoch count and early-stopping policy.

\subsection{Preprocessing and Augmentation}

Tab.~\ref{tab:sup-aug} gives the per-detector recipe. Validation and test
splits were never augmented; the multiplier applies to the training split
only.

\begin{table*}[t]
\centering\footnotesize\setlength{\tabcolsep}{5pt}
\renewcommand{\arraystretch}{1.15}
\begin{tabular}{@{}p{2.1cm}p{4.6cm}p{8.3cm}c@{}}
\toprule
Detector & Preprocessing & Augmentation & $\times N$ \\
\midrule
FDI numbering & auto-orient; fit within $768\times512$ & brightness $\pm20\%$, exposure $\pm15\%$, blur 1\,px, rotate $\pm3^\circ$ & 2 \\
Bone loss & auto-orient; fit within $768\times768$; histogram equalisation & brightness $\pm15\%$, exposure $\pm10\%$, rotate $\pm7^\circ$, horizontal flip & 2 \\
Dental history & auto-orient; stretch to $768\times448$ & brightness $\pm30\%$, exposure $\pm15\%$, blur 2\,px, noise $1.02\%$, cutout $3\times10\%$, horizontal flip & 2 \\
Jaw structure & auto-orient; fit within $576\times576$ & brightness $\pm15\%$, exposure $\pm10\%$, blur 1.5\,px & 3 \\
Periapical & auto-orient; fit within $672\times672$; histogram equalisation & brightness $\pm15\%$, exposure $\pm10\%$, blur 1\,px, rotate $\pm10^\circ$, horizontal flip & 2 \\
Dental caries & auto-orient; fit within $768\times768$ & brightness $\pm15\%$, exposure $\pm10\%$, rotate $\pm10^\circ$ & 2 \\
Impacted tooth & auto-orient; fit within $1024\times1024$ & brightness $\pm15\%$, exposure $\pm10\%$, blur 1.3\,px & 2 \\
\bottomrule
\end{tabular}
\caption{Preprocessing and augmentation per detector. $\times N$ is the
number of augmented copies generated per training image.}
\label{tab:sup-aug}
\end{table*}

Three choices in Tab.~\ref{tab:sup-aug} are deliberate rather than
default.

\paragraph{No vertical flips.} A vertically flipped panoramic radiograph
is anatomically impossible and would destroy the maxilla-mandible
distinction that every detector depends on.

\paragraph{Horizontal flip is absent from FDI numbering.} A horizontal
flip maps tooth \#11 onto tooth \#21 and would systematically corrupt
quadrant labels. It is used only where the label survives reflection:
bone loss, dental history and periapical lesions, none of which encode
laterality in the class name. Given that quadrant assignment is the
failure mode this paper addresses, augmenting the FDI detector with flips
would have undermined the component the framework depends on most.

\paragraph{Histogram equalisation on two detectors only.} Applied to bone
loss and periapical lesions, the two targets defined by low-contrast
boundaries. It is not applied elsewhere, and the tools must not re-apply
it at inference since the serving layer performs preprocessing per call.

Input resolution tracks target size, from $1024^2$ for impacted teeth
down to $576^2$ for the large jaw structures, and rotation magnitude
tracks label sensitivity, from $\pm3^\circ$ for FDI numbering where
identity is positional up to $\pm10^\circ$ for the lesion detectors.

\subsection{Tiling}
\label{sup:tiling}

The periapical and caries detectors are trained and evaluated on tiles.
For an image of width $W$ and height $H$ with $C=4$ columns and $R=2$
rows, boundaries are $x_i=\lfloor iW/C\rceil$ and
$y_j=\lfloor jH/R\rceil$. The grid is proportional rather than
fixed-size, so radiographs of differing resolution tile consistently.

Training uses the disjoint cells
$T_{ij}=[x_i,x_{i+1})\times[y_j,y_{j+1})$, which exactly partition the
image. Inference pads each cell by $\delta=128$\,px and clamps to the
image bounds, so every interior seam falls strictly inside two adjacent
tiles with $2\delta$ of shared context. A detection at $(u,v)$ in a
padded tile maps back by translation to $(u+x_0, v+y_0)$. The asymmetry
between training and inference is deliberate: disjoint training tiles
avoid teaching the model duplicate instances, while overlapping inference
tiles ensure no lesion falls on an unobserved boundary.

\paragraph{Ground truth on seams.} The two detectors handle this
differently because one is box-based and one mask-based. For periapical
lesions, a ground-truth box $b$ is clipped to the tile and retained iff
$\operatorname{area}(b\cap T_{ij})/\operatorname{area}(b)\ge0.35$ and the
clipped box has both sides at least 6\,px. A lesion divided evenly by a
seam is therefore retained in both tiles; one divided $80/20$ is retained
only in the larger fragment, and the sliver is discarded rather than
taught as a truncated example. For caries, masks are cropped per tile and
contours re-extracted on the crop, so a lesion crossing a seam becomes
two independent polygons, each retained if its area is at least
15\,px$^2$ and simplified at $\epsilon=1.0$.

\paragraph{Other settings.} Both tilers retain a $0.12$ fraction of blank
tiles as negatives, preferring hard negatives selected as the blank tiles
of highest mean intensity, which are bone-dense regions resembling
lesions rather than black background. The periapical tiler caps training
at 8{,}000 tiles and balances classes by capping majority-only tiles at
the second-largest class count, leaving validation and test
representative. The caries tiler applies a vertical flip because its
source films are stored inverted. Both use seed 42.

\clearpage
\twocolumn[
\begin{minipage}{\textwidth}
\subsection{Response time}

\centering\footnotesize\setlength{\tabcolsep}{3.5pt}
\begin{tabular}{@{}llrrrrrrrr@{}}
\toprule
& & & \multicolumn{2}{c}{LLM only} &
\multicolumn{2}{c}{Cache-free} &
\multicolumn{2}{c}{Amortized} & \\
\cmidrule(lr){4-5}\cmidrule(lr){6-7}\cmidrule(lr){8-9}
Model & Arm & $n$ & Mean & Med. & Mean & Med. & Mean & Med. & $\dagger$ \\
\midrule
\multicolumn{10}{@{}l}{\emph{Closed-ended}} \\
Gemma4-31B & Baseline   & 491 & 54.2 & 42.3 & 54.2 & 42.3 & 54.2 & 42.3 & 1 \\
Gemma4-31B & Tool-aided & 491 & 69.4 & 32.8 & 89.7 & 62.8 & 75.6 & 39.1 & 12 \\
Qwen3.5 & Baseline      & 491 & 93.8 & 80.0 & 93.8 & 80.0 & 93.8 & 80.0 & 3 \\
Qwen3.5 & Tool-aided    & 491 & 69.0 & 52.2 & 78.6 & 61.5 & 74.5 & 58.2 & 0 \\
Mistral-Large-3 & Baseline   & 491 & 15.8 & 16.1 & 15.8 & 16.1 & 15.8 & 16.1 & 0 \\
Mistral-Large-3 & Tool-aided & 491 & 18.4 & 17.6 & 26.7 & 23.9 & 23.5 & 22.9 & 0 \\
Minimax-M3 & Baseline   & 491 & 34.6 & 24.3 & 34.6 & 24.3 & 34.6 & 24.3 & 1 \\
Minimax-M3 & Tool-aided & 491 & 29.3 & 19.8 & 40.1 & 28.6 & 35.2 & 26.0 & 1 \\
\midrule
\multicolumn{10}{@{}l}{\emph{Open-ended}} \\
Gemma4-31B & Baseline   & 578 & 66.7  & 48.0 & 66.7  & 48.0 & 66.7  & 48.0 & 12 \\
Gemma4-31B & Tool-aided & 578 & 47.1  & 27.1 & 69.4  & 55.7 & 53.1  & 32.9 & 9 \\
Qwen3.5 & Baseline      & 578 & 120.1 & 92.4 & 120.1 & 92.4 & 120.1 & 92.4 & 11 \\
Qwen3.5 & Tool-aided    & 578 & 62.9  & 59.8 & 83.7  & 84.4 & 68.9  & 66.0 & 0 \\
Mistral-Large-3 & Baseline   & 578 & 24.5 & 22.0 & 24.5 & 22.0 & 24.5 & 22.0 & 0 \\
Mistral-Large-3 & Tool-aided & 578 & 27.7 & 26.0 & 46.4 & 44.0 & 33.7 & 31.9 & 0 \\
Minimax-M3 & Baseline   & 578 & 32.9 & 29.7 & 32.9 & 29.7 & 32.9 & 29.7 & 0 \\
Minimax-M3 & Tool-aided & 578 & 27.8 & 25.5 & 51.7 & 57.1 & 33.8 & 31.4 & 0 \\
\bottomrule
\end{tabular}
\captionsetup{hypcap=false}
\captionof{table}{Response time per question in seconds. \textsc{LLM} is the
measured two-call latency. \textsc{Cache-free} adds the full detector
cost to every question (cold single-question latency); \textsc{Amortized}
charges each image's detectors once and shares them across that image's
questions (the cached system in operation). The baseline arm invokes no
hosted detector, so its columns coincide. $\dagger$ counts questions
whose latency exceeds 300\,s owing to provider rate-limit backoff or connection errors
inside the timed call; medians are the robust comparison.}
\label{tab:response-time}

\vspace{4pt}
\raggedright
\subsection{Detector Inference Time Statistics}
\centering
\begin{tabular}{@{}lrrrrrrrrr@{}}
\toprule
Tool & Calls & $n$ & Mean & Med. & p95 & Min & Max & s/call & Err \\
\midrule
FDI numbering      & 1 & 98 &  2.78 &  2.38 &  4.58 & 1.43 &  10.78 & 2.78 & 1 \\
Dental history     & 1 & 98 &  1.97 &  1.63 &  3.96 & 0.74 &   5.20 & 1.97 & 1 \\
Impacted tooth     & 1 & 99 &  2.04 &  1.57 &  4.03 & 0.86 &  10.77 & 2.04 & 0 \\
Jaw structure      & 1 & 99 &  3.25 &  2.27 &  4.66 & 1.24 &  38.88 & 3.25 & 0 \\
Bone loss          & 1 & 98 &  2.78 &  2.48 &  4.87 & 1.42 &   7.35 & 2.78 & 1 \\
Periapical lesion  & 8 & 99 & 12.19 & 10.31 & 18.03 & 6.03 & 121.70 & 1.52 & 0 \\
Dental caries      & 8 & 95 &  9.43 &  9.02 & 15.08 & 5.07 &  17.47 & 1.18 & 4 \\
\midrule
All tools (summed) & 21 & -- & 34.44 & -- & -- & -- & -- & -- & -- \\
Measured           & 21 & 99 & 33.99 & 31.89 & 46.19 & 17.96 & 150.52 & -- & -- \\
\bottomrule
\end{tabular}
\captionof{table}{Per-image inference time for each detector, in seconds,
measured over $n=99$ panoramic radiographs with the evidence cache bypassed
and a warm-up image discarded. \textsc{Calls} is the number of hosted requests
one invocation issues: the periapical and caries detectors run tiled inference
over a $4\times2$ grid, so they issue eight requests per image against one for
the other five. \textsc{s/call} normalises the mean by that count, showing that
the tiled detectors are not slower per request; their per-image cost is almost
entirely the higher request count. \textsc{Err} counts failed invocations,
excluded from the statistics. \textsc{All tools} sums the per-detector means,
and \textsc{Measured} reports the observed end-to-end time to run all seven on
one image.}
\label{tab:tool-timing}
\end{minipage}
]

% =====================================================================
\section{Implementation Details}
\label{sup:implementation}
% =====================================================================

\subsection{Detection Merging}

Detections from overlapping inference tiles are merged with
class-agnostic greedy non-maximum suppression at IoU threshold
$\tau=0.4$. Class-agnostic merging is the correct choice here rather than
a default: when two overlapping tiles report the same periapical lesion
at different PAI grades, the higher-confidence grade should claim the
position outright, whereas class-aware suppression would retain both as
separate findings and report one lesion twice at conflicting severities.

\subsection{Region Queries}

Questions carrying an explicit bounding box trigger a region pass in
which the indicated area is cropped and injected into synthesis with the
detection outlined, as described by the region addendum in
Sec.~\ref{sup:prompt-synthesis}. Coordinates are handled in a single
canonical space, the absolute pixel coordinates of the full-resolution
radiograph. If the region pass returns no detection at the configured
confidence, it is retried once at half that threshold against a cached
tier. This is a detector threshold retry, not a reasoning loop: the model
is not consulted and does not re-plan.

% =====================================================================
\clearpage
\onecolumn
\section{Additional Results}
\label{sup:results}
% =====================================================================

\subsection{Full Results Under the GPT-4o Judge}

The main paper reports per-category results under the GPT-5-mini judge
and summarises the GPT-4o evaluation. Tab.~\ref{tab:sup-gpt4o} gives the
complete GPT-4o open-split breakdown. Closed-split scores are
judge-independent and identical to those in the main paper.

\begin{table*}[!ht]
\centering\small\setlength{\tabcolsep}{5pt}
\begin{tabular}{@{}lccccccc@{}}
\toprule
Model & Teeth & Patho & His & Jaw & Summ & Report & Overall \\
\midrule
gemma4:31b (baseline)      & 30.3 & 26.1 & 43.7 & 52.6 & 27.9 & 32.3 & 34.7 \\
gemma4:31b + AgenTeeth     & 48.6 & 49.2 & 55.8 & 69.7 & 36.5 & 48.9 & 51.8 \\
\addlinespace[2pt]
qwen3.5 (baseline)         & 44.4 & 33.9 & 55.2 & 68.6 & 30.6 & 38.0 & 45.1 \\
qwen3.5 + AgenTeeth        & 59.0 & 55.2 & 62.9 & 81.6 & 45.7 & 53.7 & \textbf{60.6} \\
\addlinespace[2pt]
mistral-large-3 (baseline) & 37.3 & 34.5 & 42.9 & 43.6 & 38.9 & 35.9 & 38.3 \\
mistral-large-3 + AgenTeeth& 53.1 & 57.9 & 62.1 & \textbf{83.4} & \textbf{46.7} & 55.3 & 59.0 \\
\addlinespace[2pt]
minimax-3 (baseline)       & 32.3 & 29.6 & 46.0 & 36.8 & 28.1 & 43.2 & 35.9 \\
minimax-3 + AgenTeeth      & 52.8 & 52.9 & 53.4 & 68.5 & 41.9 & \textbf{60.9} & 55.9 \\
\bottomrule
\end{tabular}
\caption{Open-ended results under the GPT-4o judge, in percent. Tool
access improves every backbone in every category without exception under
this judge. Absolute scores are lower than under GPT-5-mini while the
ordering of systems is preserved.}
\label{tab:sup-gpt4o}
\end{table*}

Under GPT-4o the improvement is unanimous: all 28 category-level
comparisons favour the tool-aided configuration, whereas under GPT-5-mini
one comparison (closed-ended Jaw for qwen3.5) does not. Absolute
open-split scores are uniformly lower under GPT-4o, consistent with a
stricter application of the same rubric rather than a different ranking
of systems.

\subsection{Closed-Ended Comparison with Published MMOral-Bench Models}
\label{sup:closed-comparison}

Tab.~\ref{tab:sup-closed} places our closed-ended results alongside the
strongest models reported in the original MMOral-Bench study
\cite{hao2025mmoral}. Closed-ended scoring is exact match and therefore
judge-independent, which is what makes this cross-paper comparison
meaningful; we do not attempt the same for the open split, where scores
depend on the judge and rubric. The published models nonetheless follow
the original MMOral zero-shot protocol while AgenTeeth uses
full-resolution images and its structured pipeline, so the comparison is
system-level rather than a controlled ablation.

\begin{table*}[!ht]
\centering\small\setlength{\tabcolsep}{5pt}
\begin{tabular}{@{}lcccccc@{}}
\toprule
Model & Teeth & Patho & His & Jaw & Summ & Overall \\
\midrule
\multicolumn{7}{@{}l}{\emph{Reported in the MMOral-Bench study}~\cite{hao2025mmoral}}\\
Ovis2-34B        & 45.8 & 51.6 & 53.9 & \textbf{79.4} & \textbf{79.2} & 56.8 \\
Qwen-QVQ-72B     & 48.7 & 49.1 & 59.3 & 74.5 & 51.0 & 56.6 \\
HealthGPT-XL32   & 39.7 & 44.1 & 51.5 & 76.4 & \textbf{79.2} & 52.0 \\
GPT-4o           & 39.7 & 41.0 & 46.7 & 55.8 & 56.3 & 45.4 \\
\midrule
\multicolumn{7}{@{}l}{\emph{Matched baselines (ours)}}\\
gemma4:31b       & 43.7 & 35.8 & 54.9 & 76.7 & 58.1 & 54.7 \\
qwen3.5          & 53.8 & 48.6 & 68.3 & 77.5 & 58.1 & 62.3 \\
mistral-large-3  & 37.7 & 35.8 & 52.1 & 49.6 & 61.3 & 40.7 \\
minimax-3        & 44.5 & 33.8 & 64.1 & 76.0 & 61.3 & 51.3 \\
\midrule
\multicolumn{7}{@{}l}{\emph{AgenTeeth (ours)}}\\
gemma4:31b       & 66.2 & 59.5 & 71.8 & 78.3 & 67.7 & 69.0 \\
qwen3.5          & \textbf{73.5} & \textbf{66.9} & \textbf{76.1} & 76.7 & 74.2 & \textbf{74.1} \\
mistral-large-3  & 60.3 & 56.1 & 67.6 & 79.1 & 71.0 & 64.4 \\
minimax-3        & 69.9 & 63.5 & \textbf{76.1} & 79.1 & 71.0 & 72.1 \\
\bottomrule
\end{tabular}
\caption{Closed-ended MMOral-Bench results, in percent. All four
AgenTeeth configurations exceed the best previously reported
closed-ended Overall (56.8), and our strongest reaches 74.1. The
exceptions are Jaw and Summary, where Ovis2-34B remains marginally
stronger; the Jaw result is consistent with the bone-loss detector
precision discussed in the main paper.}
\label{tab:sup-closed}
\end{table*}

All four configurations exceed the best previously reported closed-ended
Overall of $56.8$, by between $7.6$ and $17.3$ points. Two categories are
exceptions. Ovis2-34B remains marginally stronger on Jaw ($79.4$ against
our $79.1$) and by $5.0$ points on Summary ($79.2$ against $74.2$). The
Jaw result is the same effect analysed in the main paper: closed-ended
Jaw questions skew toward healthy findings, and our bone-loss detector's
$65.3\%$ precision propagates false positives under evidence primacy.

\subsection{Token Usage}

\begin{table}[!ht]
\centering\footnotesize\setlength{\tabcolsep}{3.5pt}
\begin{tabular}{@{}llrrrrrrrr@{}}
\toprule
Model & Arm & $n$ & In & Out & Total & Med. & Disp. & Synth. & Grand \\
\midrule
\multicolumn{10}{@{}l}{\emph{Closed-ended}} \\
Gemma4-31B & Baseline   & 491 & 3,227  & 158   & 3,385  & 3,328  & 2,118  & 1,266  & 1,661,882 \\
Gemma4-31B & Tool-aided & 491 & 5,863  & 139   & 6,003  & 4,581  & 3,856  & 2,146  & 2,947,317 \\
Qwen3.5 & Baseline      & 491 & 13,278 & 8,080 & 21,358 & 19,331 & 12,082 & 9,276  & 10,401,435 \\
Qwen3.5 & Tool-aided    & 491 & 16,629 & 4,765 & 21,394 & 19,128 & 9,582  & 11,812 & 10,504,621 \\
Mistral-Large-3 & Baseline   & 491 & 6,778  & 126   & 6,904  & 7,277  & 3,966 & 2,937  & 3,389,671 \\
Mistral-Large-3 & Tool-aided & 491 & 11,039 & 74    & 11,113 & 10,941 & 5,287 & 5,826  & 5,456,588 \\
Minimax-M3 & Baseline   & 491 & 13,838 & 3,046 & 16,884 & 15,162 & 9,294 & 7,590  & 8,290,044 \\
Minimax-M3 & Tool-aided & 491 & 18,431 & 2,069 & 20,501 & 18,454 & 9,296 & 11,204 & 10,065,772 \\
\midrule
\multicolumn{10}{@{}l}{\emph{Open-ended}} \\
Gemma4-31B & Baseline   & 578 & 2,494  & 212   & 2,707  & 2,632  & 1,603  & 1,103  & 1,564,398 \\
Gemma4-31B & Tool-aided & 578 & 4,730  & 266   & 4,996  & 5,242  & 2,661  & 2,335  & 2,887,527 \\
Qwen3.5 & Baseline      & 578 & 12,071 & 9,464 & 21,534 & 20,188 & 10,536 & 10,998 & 12,425,273 \\
Qwen3.5 & Tool-aided    & 578 & 18,110 & 3,949 & 22,058 & 22,044 & 8,598  & 13,461 & 12,749,651 \\
Mistral-Large-3 & Baseline   & 578 & 6,261  & 490   & 6,751  & 6,827  & 3,555 & 3,196  & 3,902,054 \\
Mistral-Large-3 & Tool-aided & 578 & 12,304 & 462   & 12,765 & 13,138 & 5,362 & 7,403  & 7,378,458 \\
Minimax-M3 & Baseline   & 578 & 13,666 & 2,455 & 16,121 & 15,698 & 8,551 & 7,570  & 9,317,885 \\
Minimax-M3 & Tool-aided & 578 & 20,238 & 1,837 & 22,076 & 22,083 & 9,133 & 12,942 & 12,759,680 \\
\bottomrule
\end{tabular}
\caption{Token usage per question by model and arm. Dispatch and
Synthesis are the mean total tokens of the two pipeline stages. Both
arms make exactly two model calls per question.}
\label{tab:token-usage}
\end{table}
\paragraph{Reading Table~\ref{tab:token-usage}.}
$n$ is the number of questions with recorded usage. \textsc{In} and
\textsc{Out} are the mean prompt and completion tokens per question, and
\textsc{Total} their sum; \textsc{Med.} is the median total, which is the
more robust summary given a long right tail on report-style questions.
\textsc{Disp.} and \textsc{Synth.} split that total across the two pipeline
stages, dispatch and synthesis. Because both arms issue exactly two model
calls per question, these columns are directly comparable. \textsc{Grand}
is the summed token count over the whole split, that is, the total cost of
one evaluation run.

Two patterns are visible. Tool evidence enters the prompt, so the
tool-aided arm always consumes more input tokens; the increase is the
serialised finding block and grows with the number of detectors invoked.
Against this, the tool-aided arm consistently emits \emph{fewer} output
tokens, most sharply on Qwen3.5, where completions fall from 8{,}080 to
4{,}765 on the closed split. Grounded evidence appears to shorten the
deliberation the model would otherwise perform in its own output. The two
effects partly cancel, so the net token cost of tool augmentation is
modest on the larger models and close to neutral for Qwen3.5 (21{,}358
versus 21{,}394 total tokens). It is largest in relative terms on
Gemma4-31B, whose baseline completions are already terse, leaving no
output saving to offset the added evidence.
\twocolumn

% =====================================================================
\section{Benchmark Integrity Notes}
\label{sup:benchmark}
% =====================================================================

We record three observations about MMOral-OPG-Bench encountered while
building our evaluation harness, offered as documentation for future
users of the benchmark.

\paragraph{Unanswerable closed questions.} Thirty-two of the 491
closed-ended questions have a reference answer pointing at a blank
option. Following the benchmark protocol we retain them in the
denominator, which caps achievable closed-ended accuracy at $93.5\%$ for
every system evaluated, including all published entries. The
majority-class answer alone scores $44.0\%$. Both bounds should be borne
in mind when reading closed-split numbers: an apparent score of $74.1\%$
represents $79.3\%$ of the achievable maximum.

\paragraph{Coordinate conventions.} Region-bearing questions specify
coordinates that we verified against the full-resolution image dimensions
rather than any downscaled version.

\paragraph{Image resolutions.} The released radiographs occur at three
resolutions ($2473\times1252$, $2444\times1292$, $2636\times1292$). Our
harness operates at full resolution throughout; the MMOral leaderboard
protocol downscales to 512\,px, which is one reason our baseline is not
directly comparable to published zero-shot numbers.

% =====================================================================

{\small
\bibliographystyle{ieeenat_fullname}
\bibliography{main}
}

%% file: preamble.tex
\usepackage{graphicx}
\usepackage{xcolor}
\usepackage{tikz}
\usepackage{tabularx}

\usetikzlibrary{positioning}

\definecolor{teaserblue}{HTML}{4658A6}
\definecolor{teaserbluebg}{HTML}{F3F5FC}

\definecolor{teasercoral}{HTML}{D45D63}
\definecolor{teasercoralbg}{HTML}{FFF5F4}

\definecolor{teaserteal}{HTML}{1C8C83}
\definecolor{teasertealbg}{HTML}{F1FAF8}

\definecolor{teaseramber}{HTML}{C6902F}
\definecolor{teaseramberbg}{HTML}{FFF8E8}

\definecolor{teaserink}{HTML}{293042}
\definecolor{teasermuted}{HTML}{60697A}
\definecolor{teaserborder}{HTML}{D6D9E3}

%% file: sec/0_abstract.tex
\begin{abstract}

Vision-language models (VLMs) remain largely unreliable on panoramic dental radiographs and can rely on learned anatomical priors rather than evidence in the image. This is particularly problematic for tooth localization and spatial reasoning, and fine-tuned dental VLMs can retain the same spatial biases. We present \textbf{AgenTeeth}, a model-agnostic, tool-augmented framework that grounds frozen VLMs using seven specialized dental vision experts. A question-aware orchestrator selects the relevant tools, whose detections are mapped to FDI tooth numbers or anatomical regions and returned as structured findings together with annotated image overlays. A fresh synthesis call then answers the question using this evidence, without fine-tuning the underlying VLM. On MMOral-OPG-Bench, AgenTeeth improves four backbone VLMs by 12.9--23.0 percentage points over their baselines. Our strongest configuration reaches 65.66\% on open-ended VQA, compared with 45.35\% for OralGPT-Plus. The advantage also holds at matched scale: a frozen Qwen2.5-VL-7B-Instruct with AgenTeeth reaches 48.11\%, exceeding OralGPT-Plus built on the same backbone after supervised fine-tuning and reinforcement learning for tool use. We release the framework, all seven expert models, and a dentist-annotated dataset for alveolar bone-loss detection in panoramic radiographs.

\end{abstract}

%% file: sec/Teaser_fig.tex
\begin{figure}[t]
  \centering
  \includegraphics[width=\linewidth]{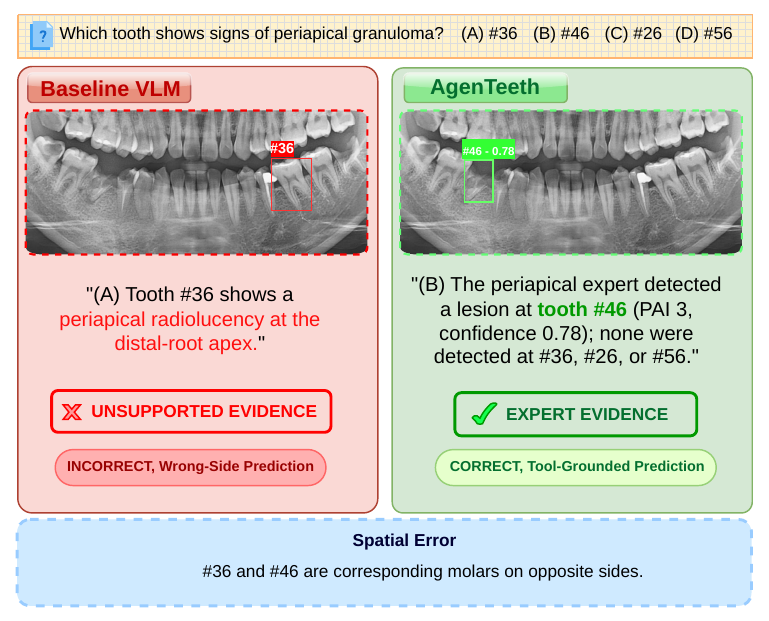}
  \caption{\textbf{AgenTeeth corrects a left--right spatial error.}
  The baseline selects tooth~\#36 and invents supporting evidence,
  whereas \textsc{AgenTeeth} grounds its answer in an expert detection
  at tooth~\#46.}
  \label{fig:teaser}
\end{figure}

%% file: sec/1_intro.tex
\section{Introduction}
\label{sec:intro}
Panoramic dental radiographs, or orthopantomograms (OPGs)~\cite{paatero1961pantomography}, provide a broad view of the dentition, both jaws, and surrounding maxillofacial structures in a single image. Clinicians read it for caries, alveolar bone
loss, periapical lesions, impacted teeth and existing restorative work,
and they must do so tooth by tooth. Reading volume is high and
specialist time is scarce, so automated support has obvious value. The
difficulty is that every finding has to be tied to a specific tooth
before it means anything clinically. This makes reliable tooth identification and spatial reasoning essential for automated panoramic X-ray analysis.

Vision-language models (VLMs) still struggle with this setting. On MMOral-Bench, a recent benchmark of 20{,}563 annotated panoramic images, the strongest of 64 evaluated models reaches only $43.31\%$ overall performance~\cite{hao2025mmoral}. Fine-tuning on dental data does not fully remove the problem, suggesting that limited domain knowledge is not the only cause. Wolf \etal~\cite{wolf2025left} show that VLMs often fail to judge the relative positions of anatomical structures in medical images. They further show that anatomical terminology can push models toward learned anatomical priors rather than the evidence visible in the image. This is especially problematic for OPGs, which are visually close to bilateral symmetry. Tooth identity in the FDI notation system~\cite{iso3950} depends directly on spatial location: each permanent tooth is represented by two digits, where the first identifies the quadrant (upper-right, upper-left, lower-left, or lower-right) and the second identifies the tooth's position from the midline. A left--right error can therefore turn an otherwise plausible observation into the wrong tooth number. We observed this failure in our baseline VLM. When asked which tooth showed signs of a periapical granuloma, it selected tooth \#36 instead of the correct tooth \#46 and supported the answer with an unsupported description of a radiolucency at the distal-root apex (Fig.~\ref{fig:teaser}). Teeth \#36 and \#46 are contralateral mandibular first molars. This example illustrates how a spatial error can produce a clinically plausible but incorrectly localized answer.

Fine-tuning does not fully remove the spatial weaknesses of general VLMs. Dental models such as OralGPT~\cite{hao2025mmoral} and DentVLM~\cite{meng2026multimodal} are built on general VLM backbones and can retain these limitations. Fine-tuning also reduces portability, since adapting to new tasks or stronger backbones may require additional training. OralGPT-Plus~\cite{fan2026oralgptplus} instead learns to use visual tools through supervised and reinforcement learning, improving visual grounding but tying the learned tool-use behavior to the trained backbone.

We take a different approach by moving domain-specific perception outside the VLM. \textbf{AgenTeeth} is a model-agnostic agentic framework built around seven specialized RF-DETR-based~\cite{robinson2026rfdetr} detection and segmentation experts for FDI tooth numbering, alveolar bone loss, jaw structures, dental history, periapical lesions, caries, and impacted teeth. The experts were developed from datasets ranging from 320 to 3{,}924 source panoramic radiographs; the small caries and periapical targets are additionally handled through tiled training. Each tool returns structured detections with class labels, confidence scores, and spatial information, together with an annotated image. Tooth-specific findings are linked to FDI numbers before being passed to the VLM. A fresh synthesis stage then answers the question using this evidence, with explicit instructions to prioritize tool findings over unsupported visual assumptions. No training is required at the reasoning layer, allowing the same framework to be used with different instruction-following VLMs. Across four backbones with matched baselines, AgenTeeth improves performance by 12.9--23.0 percentage points. We additionally evaluate Qwen2.5-VL-7B-Instruct with AgenTeeth, which reaches 48.11 on MMOral-OPG open-ended VQA. All five AgenTeeth configurations exceed the 45.35 score reported for OralGPT-Plus under the same evaluation protocol.
We make three main contributions:

\begin{itemize}

    \item We develop and release seven domain-expert vision models for panoramic dental radiography, trained on established public datasets ranging from 320 to 3{,}924 source images.

    \item We introduce and openly release a new alveolar bone-loss object-detection dataset built from 943 TUFTS~\cite{panetta2022tufts} panoramic radiographs, with all images manually annotated by a dentist.

    \item We propose \textbf{AgenTeeth}, a model-agnostic, tool-augmented agentic framework for panoramic dental VQA that consistently improves multiple VLM backbones and surpasses prior systems on MMOral-OPG-Bench.
    
\end{itemize}
Upon acceptance, we will release AgenTeeth, all seven trained tools, and the
dentist-annotated bone-loss dataset.

%% file: sec/2_relatedwork.tex
\section{Related Work}
\label{sec:related}

\subsection{VLMs for Dental Radiography}

VLMs have increasingly been adapted to medical imaging, but
fine-grained grounding between image regions and clinical concepts
remains challenging~\cite{lin2025taming}. Panoramic dental radiographs are particularly challenging because a single image contains
the full dentition and surrounding anatomy, with substantial structural overlap
and many small pathological findings.

MMOral~\cite{hao2025mmoral} introduced the first large-scale multimodal
instruction dataset and benchmark for panoramic dental X-rays. It contains
20{,}563 annotated images and 1.3 million instruction-following instances, with
MMOral-Bench evaluating five diagnostic dimensions. Among 64 large vision-language models (LVLMs) evaluated in
the study, the best-performing model, GPT-4o, achieved only
43.31\% overall performance.
The same work introduced OralGPT, obtained by supervised fine-tuning
Qwen2.5-VL-7B on MMOral. DentVLM~\cite{meng2026multimodal} similarly adapts VLMs
to dental imaging across multiple modalities.

Fine-tuning on dental data does not by itself guarantee reliable visual grounding. 
Wolf \etal~\cite{wolf2025left} show that some VLMs, particularly GPT-4o and 
Pixtral, favor learned anatomical priors when those priors conflict with the 
provided image. When anatomical names were removed and structures were indicated 
only by visual markers, both models improved substantially, suggesting that 
explicit visual grounding can reduce this bias. 

\subsection{Tool-Augmented and Agentic Medical VLMs}

A growing line of work augments VLMs with specialized agents
and visual tools. VipAct~\cite{zhang2026vipact} combines an
orchestrator with specialized agents and vision experts for
fine-grained perception. AIDE~\cite{chiu2025aide} instead uses
domain experts to refine training examples without relying on a
stronger teacher model. Related agentic approaches have also
emerged in medicine. MMedAgent~\cite{li2024mmedagent} learns
to select among specialized medical tools across multiple
modalities, while MedRAX~\cite{fallahpour2025medrax} dynamically
invokes expert tools for chest X-ray analysis. Other medical
multi-agent systems include MDAgents~\cite{kim2024mdagents}
and MedAgent-Pro~\cite{wang2026medagentpro}.

Agentic dental systems have emerged more recently. OralAgent
\cite{hao2026oralagent} combines 22 visual tools with retrieval over a
large bilingual dental corpus and uses a ReAct-style workflow for multimodal
dental analysis. OPGAgent~\cite{yu2026opgagent} focuses specifically on
panoramic radiographs. It gathers evidence hierarchically at the global,
quadrant, and tooth levels, then resolves disagreements using a consensus
subagent constrained by anatomical information. OralGPT-Plus
\cite{fan2026oralgptplus} follows a different strategy. It trains
Qwen2.5-VL models to use localized inspection and contralateral comparison
through supervised fine-tuning and reinforcement learning, which enables iterative
and symmetry-aware reinspection.

\paragraph{Relation to our work.}
AgenTeeth differs from these systems in where domain expertise is placed and
how evidence is used. The reasoning VLM remains frozen, while dental perception
is delegated to seven external domain experts. The framework therefore does not
require SFT or reinforcement learning when the orchestrator backbone is changed.
Unlike systems whose tools mainly provide additional views or whose
orchestrator arbitrates between tool output and its own perception, our tools
return explicit clinical detections together with annotated overlays. These
findings are grounded to FDI tooth numbers or anatomical regions before
synthesis, and the orchestrator is instructed to prioritize this evidence over
unsupported visual assumptions. This design separates perception from
reasoning and allows the same tool set to be used across different VLM
backbones.

%% file: sec/3_domainexpert.tex
\section{Domain Expert Tools}
\label{sec:domainexpert}

AgenTeeth uses seven specialized vision models to provide external evidence for
panoramic dental VQA. All experts use RF-DETR Medium
\cite{robinson2026rfdetr} and are trained and deployed through
Roboflow~\cite{roboflowdeploy}. Four perform object detection and three perform
instance segmentation. Their datasets and source-image counts are provided in Supplementary Sec. C.

\subsection{Tool Interface}
\label{sec:contract}

All experts share a common interface, returning each detection's location,
class label, confidence score, and an annotated radiograph. Tooth-specific
findings are linked to FDI numbers before reaching the VLM, while anatomical
structures such as the mandibular canal or maxillary sinus retain their labels.
The value of the annotated overlays is evaluated in Sec.~\ref{sec:ablation}.

\subsection{Training Data}

All seven experts are initialized from pretrained RF-DETR weights and trained on
public panoramic-radiograph datasets, with additional annotations where suitable
labels were unavailable. The bone-loss dataset contains 943 TUFTS~\cite{panetta2022tufts} radiographs,
all manually annotated by a dentist; we also add TUFTS annotations for dental
history and PDX annotations for jaw structures.

For FDI numbering, we merge TUFTS~\cite{panetta2022tufts},
DUAL~\cite{zhou2024dual}, DENTEX~\cite{hamamci2023dentex}, and
PDX~\cite{brahmi2025pdx}. TUFTS labels are converted from Universal to FDI
notation, and pediatric samples are excluded to retain the 32 permanent tooth
classes. Training, validation, and test partitions are preserved when datasets
are merged, and augmentation is applied only to training data. Full per-tool
dataset sources, image counts, and split details are provided in
Supplementary Sec.~C.

\subsection{Full-Image FDI Grounding}

FDI numbering requires information from the complete dental arch. In the FDI
system, the first digit identifies the quadrant and the second identifies the
tooth position from the midline. Local appearance alone is therefore
insufficient: a cropped first molar may correspond to \#16, \#26, \#36, or
\#46 depending on its position in the full radiograph. We consequently run the
FDI expert on the complete panorama and do not tile it.

This tool provides the spatial anchor for the remaining experts. Tooth-specific
detections are associated with FDI detections before they reach the VLM, so an
FDI error can propagate to otherwise correct pathology detections. Our ablation
in Sec.~\ref{sec:ablation} supports this dependence: replacing the FDI expert
from 92.4 to 88.1 mAP@50 reduces average downstream VQA performance by
3.32 percentage points.
\subsection{Tiled Lesion Detection}

Dental caries and periapical lesions occupy only a small fraction of a panoramic
radiograph. We therefore train these two experts on a proportional
$4\times2$ grid. For an image of width $W$ and height $H$, the grid boundaries
are

\begin{equation}
x_i = \left\lfloor \frac{iW}{4} \right\rceil,\quad
y_j = \left\lfloor \frac{jH}{2} \right\rceil,
\label{eq:tile-grid}
\end{equation}

where $i=0,\ldots,4$ and $j=0,\ldots,2$. Training tiles are non-overlapping,
and the dataset is split at the panoramic-image level before tiling so that
tiles from the same radiograph never appear across different data splits.

At inference, each tile is padded by $\delta=128$ pixels to preserve context
near tile boundaries:

\begin{equation}
\begin{aligned}
T^{\mathrm{inf}}_{ij}
&= [\max(0,x_i-\delta),\,\min(W,x_{i+1}+\delta)] \\
&\quad\times
   [\max(0,y_j-\delta),\,\min(H,y_{j+1}+\delta)] .
\end{aligned}
\label{eq:tile-inference}
\end{equation}
Predictions are mapped back to full-image coordinates and duplicate detections
from overlapping tiles are removed using class-agnostic non-maximum suppression
(NMS)~\cite{neubeck2006efficient} at an IoU threshold of $0.4$. The final annotated radiograph is then reconstructed before being
passed to the orchestrator. Further details on boundary handling and annotation
processing are provided in Supplementary Sec.~B.
\subsection{Tool Performance}

Tab.~\ref{tab:toolperf} reports the held-out performance of the seven experts.
Performance varies across tasks. Jaw-structure and FDI detection are the
strongest, reaching 97.9 and 92.4 mAP@50, followed by impacted-tooth detection
at 91.0 and dental history at 87.6. The lesion-focused tools are more
challenging: periapical lesions reach 66.5 mAP@50, while bone loss and caries
reach 57.5 and 54.6, respectively.

These differences are important because AgenTeeth gives tool evidence priority
during synthesis, so detector errors can propagate to the final answer. Bone
loss is particularly challenging because its annotations are strongly
imbalanced toward healthy bone; the detector reaches 65.3\% precision and
52.1\% recall. We examine the effect of tool quality directly in
Sec.~\ref{sec:ablation} and discuss the remaining failure modes in
Sec.~\ref{sec:limitations}.

% =====================================================================
% TABLE 2: expert performance
% =====================================================================

\begin{table}[t]
\centering
\small
\setlength{\tabcolsep}{4.5pt}
\begin{tabular}{@{}lcccc@{}}
\toprule
Tool & mAP@50 & Prec. & Rec. & F1 \\
\midrule
Jaw structure     & 97.9 & 98.3 & 98.3 & 98.3 \\
FDI numbering     & 92.4 & 91.0 & 90.1 & 90.5 \\
Impacted tooth    & 91.0 & 74.7 & 83.8 & 79.0 \\
Dental history    & 87.6 & 89.0 & 80.4 & 84.4 \\
Periapical lesion & 66.5 & 59.4 & 65.0 & 62.1 \\
Bone loss         & 57.5 & 65.3 & 52.1 & 58.0 \\
Dental caries     & 54.6 & 64.7 & 51.5 & 57.4 \\
\bottomrule
\end{tabular}

\caption{Held-out performance of the domain experts, reported in percent.}

\label{tab:toolperf}
\end{table}

%% file: sec/4_framework.tex
\section{Framework}
\label{sec:framework}

AgenTeeth answers each panoramic-radiograph question in two passes over the same
frozen orchestrator VLM. The first pass selects which trained domain experts to
consult, while the second synthesizes the final answer from their outputs. A
deterministic grounding layer between the two converts detector geometry into
tooth-numbered or anatomically localized findings. Fig.~\ref{fig:framework}
summarizes the full pipeline.

Only the orchestrator VLM remains untrained and interchangeable; the seven
specialized dental tools are trained separately. The same off-the-shelf VLM is
used for both dispatch and synthesis, allowing the tool suite to be paired with
different instruction-following backbones without retraining the reasoning
layer. The full dispatch and synthesis prompts are provided in
Supplementary Sec.~A.
\subsection{Dispatch}

The first pass reads the question and selects tools. Dispatch is
governed by three rules, shown as the dispatch policy in
Fig.~\ref{fig:framework}.

The FDI numbering tool is always invoked because tooth identity provides the
grounding reference for tooth-specific findings. The remaining six experts are
selected from the question text; for example, restoration questions trigger the
dental-history tool, while impaction questions trigger the impacted-tooth tool.
If the dispatcher is uncertain, it invokes all remaining experts to avoid
missing relevant evidence.

Dispatch uses only the question text. Although MMOral-OPG-Bench provides
category labels, AgenTeeth does not use them, preventing evaluation metadata
from leaking into tool selection.
\subsection{Grounding}

Detectors return boxes or masks, but the final answer needs findings linked to
specific teeth. The grounding layer performs this association deterministically.
Tooth numbers themselves come from the trained FDI detector; the grounding
logic only decides which detected FDI tooth, if any, a finding belongs to.

For each finding, we first check whether its centre lies inside an FDI tooth box.
If so, the finding is assigned to that tooth. Otherwise, we match it to the
nearest tooth within a distance of one tooth-box diagonal. For periapical
lesions, this limit is increased to $1.8\times$ because the lesion centre often
lies near the root apex, outside the main tooth box. Findings that match neither
rule are marked \emph{unbound} rather than being assigned a guessed tooth number.

This design removes tooth attribution from the VLM. The model receives
tooth-grounded findings produced from FDI detections instead of inferring
laterality or tooth identity from the radiograph itself. Quadrants are taken
directly from the FDI labels, which reduces reliance on the spatial judgements
that VLMs are known to handle poorly~\cite{wolf2025left}. Errors can still arise
if the FDI detector itself predicts the wrong tooth, which we discuss in
Sec.~\ref{sec:limitations}.

\subsection{The Structured Evidence Packet}

Grounded detections are assembled into a tooth-level evidence packet, for
example, \texttt{\#36 Root\_Canal\_treated\_tooth (0.88)
[Dental\_History]}. Findings that are not tooth-specific, such as the mandibular
canal or maxillary sinus, are listed separately as anatomical structures.

The synthesis prompt does not include raw coordinates or polygons. Instead, it
receives concise clinical findings with confidence scores, while the original
geometry is stored separately to preserve an audit trail back to each
detection.

The second pass also receives visual evidence: the original radiograph,
relevant annotated overlays, and, when available, a crop of the queried region.
We cap the visual context at four images to balance evidence coverage with model
attention; adding too many similar views can dilute focus on the most relevant
findings. The selected images therefore prioritize the raw OPG and the overlays
most closely related to the question. Sec.~\ref{sec:ablation} evaluates the
benefit of these annotated images beyond the structured text alone.

\subsection{Synthesis Under Evidence Primacy}

The second pass is a fresh call to the same VLM, with no tool access or dispatch
history. It receives only the question, the radiograph, and the curated evidence
packet. This prevents the synthesis stage from being influenced by the
dispatcher's intermediate reasoning or from re-running tools.

During synthesis, the VLM is instructed to prioritize tool evidence over its own
visual impression.
However, this trust is constrained by detector-specific guardrails. Unbound
findings remain explicitly marked and cannot be assigned to a tooth. The
periapical and caries tools may occasionally produce detections in anatomically
implausible regions, such as areas far from a tooth or root apex; the prompt
therefore allows the VLM to reject such findings when their location is
clinically impossible. A separate caveat is used for the jaw-structure tool,
which does not include a nasal-cavity class, so the absence of a detection is
not treated as evidence that the structure is absent. These guardrails preserve
tool primacy while accounting for known failure modes of individual experts.

\subsection{Output}
Closed-ended questions are answered with an option letter in a fixed terminal
format for exact-match parsing. Open-ended questions are answered in prose, with
tooth-specific findings referenced by FDI number.

For report-style questions, we prompt the model to follow the report structure
used by MMOral-OPG-Bench, whose reference reports were organized around the
components expected in a dental report~\cite{hao2025mmoral}. We apply the same
report-format instruction to both experimental conditions: the \emph{tool-aided}
AgenTeeth setting and the matched \emph{baseline} setting without domain-expert
evidence. This keeps response format consistent, so differences in score are more
likely to reflect answer content rather than presentation.

\subsection{Model-Agnostic Deployment}

Because the orchestrator VLM is not fine-tuned, it remains interchangeable. The seven domain experts are trained separately and exposed through a fixed tool interface, so the same reasoning pipeline can be paired with different instruction-following VLM backbones without retraining the orchestrator.

 In the tool-aided setting, the orchestrator receives evidence from the seven trained experts; in the matched baseline, those experts are removed while the remaining pipeline is kept unchanged. This isolates the contribution of external tool evidence more directly than approaches in which tool-use behavior is learned inside the model weights.

%% file: sec/5_experimental_setup.tex
\section{Experimental Setup}
\label{sec:experimental-setup}

We evaluate AgenTeeth on MMOral-OPG-Bench~\cite{hao2025mmoral}, containing
491 closed-ended and 578 open-ended questions over 100 panoramic radiographs.
Our main controlled experiment compares each tool-aided AgenTeeth configuration
against a matched baseline using the same VLM, prompts, two-pass pipeline, and
evaluation procedure, but without the seven domain-expert tools.

We evaluate four VLM backbones in this paired setting and additionally test
Qwen2.5-VL-7B-Instruct with AgenTeeth for direct comparison with
OralGPT-Plus~\cite{fan2026oralgptplus}. Closed-ended questions are scored using
the official MMOral exact-match evaluation, while open-ended responses are
evaluated with the OralGPT-Plus evaluation protocol using GPT-5-mini as judge.
Full benchmark details, model serving, detector inference and caching,
decoding settings, evaluation procedure, and implementation details are provided
in Supplementary Sec.~B, while inference-time measurements for the tools,
AgenTeeth, and the matched baseline are reported in Supplementary Sec.~D.

%% file: sec/6_results.tex
\section{Results}
\label{sec:results}
\subsection{Main Results}

Tab.~\ref{tab:main} reports matched baseline and tool-aided results for four
backbones, together with published MMOral-OPG results. AgenTeeth improves every
backbone on both splits: closed-ended Overall increases by 11.8--23.7 points and
open-ended Overall by 14.0--22.4 points, with mean gains of 12.9--23.0 points.
Of the 44 category-level comparisons, 43 favor the tool-aided setting; the only
reversal is closed-ended Jaw for qwen3.5.

The gains are consistent across backbones. All four AgenTeeth configurations
also exceed the strongest closed-ended Overall reported by MMOral, Ovis2-34B at
56.8\%, reaching 64.4--74.1\%. On the open split, our strongest configuration,
minimax-3, reaches 65.7\%, compared with 45.4\% for OralGPT-Plus-7B and 42.3\%
for GPT-5.

\noindent
\textbf{Parameter-matched comparison:}
To control for model scale, we also evaluate AgenTeeth with
Qwen2.5-VL-7B-Instruct, the same backbone used by OralGPT-Plus. With the
orchestrator kept frozen, AgenTeeth reaches 48.1 open-ended Overall, compared
with 45.4 for OralGPT-Plus-7B after supervised fine-tuning and reinforcement
learning. The largest gains appear in Pathology, Summary, and History, while
OralGPT-Plus remains stronger on Jaw and slightly better on Teeth. The Jaw gap
likely reflects its explicit \texttt{Mirror-In} symmetry mechanism, which
AgenTeeth does not include (Sec.~\ref{sec:limitations}).

\noindent
\textbf{Comparison with MMOral-Bench models:}
Closed-ended evaluation is exact-match and judge-independent, allowing direct
comparison with the 64 LVLMs evaluated by MMOral~\cite{hao2025mmoral}. Among
the 36 models reported in the main text of MMOral, the strongest closed-ended
Overall score is 51.53\% from HealthGPT-XL32. All four AgenTeeth
configurations exceed this result, reaching 64.4--74.1\%, with qwen3.5
achieving the best score of 74.1\% (+22.6 points). A full comparison with
the MMOral models, including the complete 64-model ranking, is provided in
the supplementary material.

\noindent
\textbf{Judge Robustness:}
Prior dental VQA studies use different LLM judges: OralGPT uses GPT-4-turbo,
whereas OralGPT-Plus uses GPT-5-mini. Since GPT-4-turbo is deprecated, we use
GPT-4o as a current alternative and GPT-5-mini to match the OralGPT-Plus
evaluation setting. AgenTeeth improves every backbone under both judges:
15.5--20.7 points with GPT-4o and 14.0--22.4 points with GPT-5-mini. Although
absolute scores and model ranking vary with the judge, the consistent gains
show that the benefit of AgenTeeth is not tied to a particular judge.

\subsection{Where Tools Do Not Help: Closed-Ended Jaw}

Closed-ended Jaw is the only category where tool gains are negligible or
slightly negative for some backbones. This appears to come from two effects.
First, several unaided models already score unusually high on this category,
likely because many questions favor healthy findings. Second, the bone-loss
expert has limited precision (65.3\%), so false positives can be propagated when
tool evidence is prioritized.

This effect is much smaller on open-ended Jaw, where the same backbones gain
8.7--36.9 points, and mistral-large-3 improves by 29.5 points from a lower
closed-Jaw baseline. We therefore treat this as a limitation of tool quality
rather than of the framework itself, and test that interpretation in
Sec.~\ref{sec:ablation}.

%% file: sec/7_ablation_studies.tex
% ---------------------------------------------------------------------
% TABLE 5, tool-quality swap. mistral-large-3, GPT-5-mini judge.
% Full width: the Jaw column carrying 0.0 in both splits is the point of
% the table, so it must be visible alongside everything else.
% ---------------------------------------------------------------------
\begin{table*}[t]
\centering\footnotesize\setlength{\tabcolsep}{2.9pt}
\begin{tabular}{@{}lccccccc@{\hskip 6pt}ccccccc@{}}
\toprule
& \multicolumn{1}{c}{Tool} & \multicolumn{6}{c}{Closed-ended VQA} & \multicolumn{7}{c}{Open-ended VQA} \\
\cmidrule(lr){2-2}\cmidrule(lr){3-8}\cmidrule(l){9-15}
Config. & mAP@50 & Teeth & Patho & His & Jaw & Summ & Overall
              & Teeth & Patho & His & Jaw & Summ & Report & Overall \\
\midrule
AgenTeeth (full)     & --   & 60.3 & 56.1 & 67.6 & 79.1 & 71.0 & 64.4
                     & 57.9 & 57.7 & 69.1 & 89.1 & 65.7 & 39.9 & 61.1 \\
\midrule
\;\;w/ weaker FDI            & $92.4\!\rightarrow\!88.1$
                     & 58.3 & 54.1 & 64.8 & 79.1 & 64.5 & 62.9
                     & 49.8 & 54.0 & 60.3 & 89.1 & 63.6 & 35.8 & 56.0 \\
\;\;\emph{difference}        & $-4.3$
                     & \textit{--2.0} & \textit{--2.0} & \textit{--2.8} & \textbf{\textit{0.0}} & \textit{--6.5} & \textit{--1.5}
                     & \textit{--8.0} & \textit{--3.6} & \textit{--8.8} & \textbf{\textit{0.0}} & \textit{--2.1} & \textit{--4.1} & \textit{--5.1} \\
\midrule
\;\;w/ weaker history & $87.6\!\rightarrow\!59.0$
                     & 58.3 & 56.1 & 60.6 & 77.5 & 71.0 & 62.3
                     & 53.7 & 56.3 & 56.2 & 88.6 & 64.0 & 36.5 & 57.8 \\
\;\;\emph{difference}        & $-28.6$
                     & \textit{--2.0} & \textit{0.0} & \textbf{\textit{--7.0}} & \textit{--1.6} & \textit{0.0} & \textit{--2.1}
                     & \textit{--4.2} & \textit{--1.4} & \textbf{\textit{--12.9}} & \textit{--0.5} & \textit{--1.7} & \textit{--3.4} & \textit{--3.4} \\
\bottomrule
\end{tabular}
\caption{\textbf{Framework accuracy tracks detector accuracy.} Replacing either
the FDI or dental-history expert with a weaker version reduces downstream VQA
performance while all other components remain fixed. Results use
mistral-large-3 as the backbone.}
\label{tab:toolswap}
\end{table*}
%=====================================================================
%  TABLES for Sec. Ablation Studies
% =====================================================================

% ---------------------------------------------------------------------
% TABLE 6, annotated overlay ablation. mistral-large-3, GPT-5-mini judge.
% ---------------------------------------------------------------------
\begin{table}[t]
\centering\small\setlength{\tabcolsep}{4pt}
\begin{tabular}{@{}lcccccc@{}}
\toprule
& \multicolumn{2}{c}{Closed} & \multicolumn{3}{c}{Open} & \\
\cmidrule(lr){2-3}\cmidrule(lr){4-6}
Images & Jaw & Overall & His & Summ & Overall & Avg \\
\midrule
1 & 78.3 & \textbf{64.6} & 64.4 & 59.4 & 58.5 & 61.5 \\
4 & \textbf{79.1} & 64.4 & \textbf{69.1} & \textbf{65.7} & \textbf{61.1} & \textbf{62.8} \\
\bottomrule
\end{tabular}
\caption{\textbf{Effect of annotated overlays.} Increasing synthesis input from
one to four overlays improves open-ended performance by 2.6 points, with
negligible change on the closed split. Results use mistral-large-3.}
\label{tab:imgablation}
\end{table}

\section{Ablation Studies}
\label{sec:ablation}

We ablate three components: the detectors themselves, the accuracy of an
individual detector, and the annotated overlays returned alongside the
structured findings.

\subsection{Contribution of the Detectors}

The matched baseline comparison serves as our primary ablation: the two settings
differ only in access to the seven domain experts. Removing them reduces
closed-ended performance by 11.8--23.7 points and open-ended performance by
14.0--22.4 points across four backbones.

The gain is largest for the weakest unaided model, mistral-large-3, and smallest
for the strongest, qwen3.5, suggesting that external expert evidence is most
valuable when the base VLM's own visual perception is limited.

\subsection{Tool Quality Ablation}
\label{sec:ablation-toolquality}

To test whether framework performance depends on detector quality, we replace
one expert at a time with a weaker version while keeping the rest of the pipeline
fixed. Using mistral-large-3, the FDI tool is reduced from 92.4 to 88.1
mAP@50, while the dental-history tool is reduced from 87.6 to 59.0 mAP@50.

Both swaps reduce downstream VQA performance
(Tab.~\ref{tab:toolswap}). Weakening dental history lowers mean performance by
2.7 points, with the largest drops in the History category: $-7.0$ closed-ended
and $-12.9$ open-ended. In contrast, weakening FDI lowers mean performance by
3.3 points across several tooth-dependent categories, including $-8.0$ in
open-ended Teeth and $-8.8$ in History.

Jaw remains unchanged under the FDI swap ($79.1$ closed-ended and $89.1$
open-ended), because jaw structures are not assigned through FDI binding. This
acts as an internal control and supports the expected dependency between tool
quality and the categories that consume its evidence.

The FDI result also shows that attribution quality has disproportionate impact:
a 4.3-point drop in FDI mAP@50 causes a larger framework loss than the 28.6-point
drop in dental-history mAP@50. This suggests that improving the grounding tool
may yield greater downstream benefit than improving a task-specific detector.
Overall, the results indicate that lower-performing experts represent
recoverable headroom rather than a fixed limitation of the framework.
\subsection{Contribution of Annotated Overlays}

To isolate the value of visual overlays, we vary only the number of annotated
images passed to synthesis while keeping the structured textual evidence fixed.
With mistral-large-3, increasing the visual context from one to four overlays
raises open-ended Overall from 58.5 to 61.1, while closed-ended performance
remains essentially unchanged (Tab.~\ref{tab:imgablation}).
The gain is modest compared with the full detector contribution, but it shows
that overlays provide complementary information beyond structured findings.
Their benefit is concentrated in open-ended VQA, where describing extent and
spatial relations can require visual context that a short textual evidence
packet does not fully capture.

%% file: sec/8_limitations.tex
\section{Limitations}
\label{sec:limitations}
 
\textbf{Detector quality bounds the framework.}
AgenTeeth inherits the limitations of its domain experts. Bone loss,
dental caries, and periapical lesion detection achieve mAP@50 of
$57.5$, $54.6$, and $66.5$, respectively
(Tab.~\ref{tab:toolperf}), below the other four experts
($87.6$--$97.9$). Caries and periapical lesions are small,
low-contrast targets for which tiling helps but does not fully resolve
the detection difficulty. Bone-loss labels are also highly imbalanced,
with healthy annotations outnumbering severe cases by more than
$100{:}1$, contributing to its lower recall ($52.1\%$). Our ablation
in Sec.~\ref{sec:ablation-toolquality} further shows that framework
performance improves with detector quality, suggesting that stronger
experts provide clear headroom.

\noindent
\textbf{Tool errors can propagate to the final answer.}
The orchestrator is instructed to prioritize tool evidence over its own
visual judgement. This reduces reliance on unreliable VLM perception,
but also exposes the system to detector errors. For example, the bone
loss expert has $65.3\%$ precision, leaving substantial false-positive
risk and likely contributing to the weaker gains observed on some
closed-ended Jaw questions (Sec.~\ref{sec:results}). The trade-off is
also backbone-dependent: some VLMs recognize implausible detections
from the structured evidence or annotated overlay and correct them,
whereas others propagate them. Evidence primacy is therefore encouraged
through prompting rather than guaranteed by construction.

\noindent
\textbf{No explicit symmetry or iterative verification.}
AgenTeeth does not perform contralateral comparison, i.e.,
explicitly comparing tooth \#16 with \#26. This remains an important
capability gap for panoramic radiographs. This is especially relevant given the
benefit of symmetry-aware inspection reported by OralGPT-Plus.
The framework also uses a fixed two-pass pipeline, so the synthesis
stage cannot re-invoke tools or request additional inspection. For the small subset of region-specific questions, a limited fallback
retries the relevant detector at half its original confidence threshold
when no suitable detection is found. This is a detector-level retry
rather than a model-driven verification loop.

\noindent
\textbf{Evaluation caveats.}
Each configuration was evaluated in a single run, so stochastic
variation across repeated runs is not measured. In addition, our
baseline is not identical to the MMOral zero-shot leaderboard protocol:
it uses full-resolution images and our structured prompting pipeline.
Published leaderboard comparisons should therefore be interpreted as
system-level comparisons, while the internal tool-versus-baseline
ablation provides the controlled estimate of tool contribution
(Sec.~\ref{sec:experimental-setup}). Finally, 32 closed-ended questions
have invalid reference options, limiting the maximum achievable closed
accuracy to $93.5\%$ for every evaluated system.

%% file: sec/9_conclusion.tex
\section{Conclusion}
\label{sec:conclusion}

We presented AgenTeeth, a model-agnostic framework that moves
domain-specific perception outside the VLM. Seven specialized detectors
provide dental findings, which are deterministically grounded to FDI
tooth numbers and passed to a frozen orchestrator for evidence-based
reasoning. No training is required at the reasoning layer, and the
framework can incorporate additional domain experts as task requirements
change. Across four backbones, AgenTeeth improves accuracy by
$12.9$--$23.0$ percentage points over the corresponding detector-free
baselines, and all configurations exceed the best previously reported
result on MMOral-OPG open-ended VQA. At matched scale, AgenTeeth with a
frozen Qwen2.5-VL-7B-Instruct also outperforms the same backbone trained
for tool use through supervised fine-tuning and reinforcement learning.
The supplementary comparison further shows strong closed-ended
performance against models evaluated in MMOral-Bench. Future work will focus on stronger
small-lesion detectors and explicit contralateral comparison, addressing
the two main limitations identified by our experiments.